\documentclass{article}

\usepackage{hyperref}

\usepackage[accepted]{icml2024}

\usepackage{amsmath}
\usepackage{amssymb}
\usepackage{mathtools}
\usepackage{amsthm}
\usepackage{hyperref}
\usepackage{amsmath,flushend}
\usepackage{booktabs}
\usepackage{amsfonts}
\usepackage{amsbsy}
\usepackage{bbm}
\usepackage{bm}
\usepackage{enumitem}
\usepackage{lipsum}
\usepackage{multirow}
\usepackage[normalem]{ulem}
\usepackage{url}

\usepackage{graphicx}
\usepackage{subcaption}
\usepackage{graphicx,color}

\usepackage{wrapfig, flushend, multirow}
\usepackage{longtable}

\usepackage{graphicx}
\usepackage{colortbl}

\newcommand{\name}{TacSas}

\usepackage[capitalize,noabbrev]{cleveref}

\theoremstyle{plain}
\newtheorem{theorem}{Theorem}[section]

\newtheorem{lemma}[theorem]{Lemma}

\theoremstyle{definition}

\theoremstyle{remark}
\newtheorem{remark}[theorem]{Remark}

\usepackage[textsize=tiny]{todonotes}

\icmltitlerunning{Generating Fine-Grained Causality in Climate Time Series Data for Forecasting and Anomaly Detection}

\begin{document}

\twocolumn[
\icmltitle{Generating Fine-Grained Causality in Climate Time Series Data\\ for Forecasting and Anomaly Detection}



\icmlsetsymbol{equal}{*}

\begin{icmlauthorlist}
\icmlauthor{Dongqi Fu}{uiuc}
\icmlauthor{Yada Zhu}{ibm,mitibm}
\icmlauthor{Hanghang Tong}{uiuc}
\icmlauthor{Kommy Weldemariam}{amazon}
\icmlauthor{Onkar Bhardwaj}{mitibm}
\icmlauthor{Jingrui He}{uiuc}
\end{icmlauthorlist}

\icmlaffiliation{uiuc}{University of Illinois Urbana-Champaign, USA}
\icmlaffiliation{ibm}{IBM Research, USA}
\icmlaffiliation{mitibm}{MIT-IBM Watson AI Lab, USA}
\icmlaffiliation{amazon}{Amazon Sustainability Science and Innovation, USA}

\icmlcorrespondingauthor{Jingrui He}{jingrui@illinois.edu}

\icmlkeywords{Machine Learning, ICML}

\vskip 0.3in
]



\printAffiliationsAndNotice{}  

\begin{abstract}
Understanding the causal interaction of time series variables can contribute to time series data analysis for many real-world applications, such as climate forecasting and extreme weather alerts.
However, causal relationships are difficult to be fully observed in real-world complex settings, such as spatial-temporal data from deployed sensor networks.
Therefore, to capture fine-grained causal relations among spatial-temporal variables for further a more accurate and reliable time series analysis, we first design a conceptual fine-grained causal model named \textit{TBN Granger Causality}, which adds time-respecting Bayesian Networks to the previous time-lagged Neural Granger Causality to offset the instantaneous effects.
Second, we propose an end-to-end deep generative model called \textit{\name}, which discovers TBN Granger Causality in a generative manner to help forecast time series data and detect possible anomalies during the forecast.
For evaluations, besides the causality discovery benchmark Lorenz-96, we also test \name\ on climate benchmark ERA5 for climate forecasting and the extreme weather benchmark of NOAA for extreme weather alerts.
\end{abstract}

\section{Introduction}
``\textit{Climate science investigates the structure and dynamics of earth’s climate system and seeks to understand how global, regional, and local climates are maintained as well as the processes by which they change over time}",\footnote{\url{https://plato.stanford.edu/entries/climate-science/}} where the corresponding data are usually stored in the format of time series, recording the climate features, geo-locations, time attributes, etc.

In time series data, variables often exhibit high-dimensional characteristics, and correlation between variables tends to be intricate, hard to obtain, and encompassing aspects such as non-linearity and time dependency. Taking the climate time series data as an example, multiple variables such as temperature, wind gust, atmospheric water content, and solar radiation co-appear on the time axis. Although we can access their tabular representations, their interactions are typically complex (e.g., non-linear, time-dependent), making it difficult to understand and capture the time series evolution trend and latent distribution of values. As a result, this complexity may lead to sub-optimal performance in time series analysis, such as time series forecasting and anomaly detection. 

Structure learning has recently gained much attention, such as~\cite{DBLP:conf/iclr/LiYS018, DBLP:conf/kdd/WuPL0CZ20, DBLP:conf/icdm/ZhaoWDHCTXBTZ20, DBLP:conf/nips/CaoWDZZHTXBTZ20, DBLP:conf/iclr/Shang0B21, DBLP:conf/aaai/DengH21, DBLP:conf/iclr/MarcinkevicsV21, DBLP:journals/corr/abs-2202-02195, DBLP:journals/pami/TankCFSF22, DBLP:journals/pami/SpadonHBMRS22, DBLP:journals/fdata/FuH22, DBLP:conf/cikm/ZhouZF0H22, gong2023rhino, DBLP:conf/wsdm/FuXTH23, DBLP:conf/www/LiFH23, DBLP:journals/corr/abs-2403-16030}. Among others, causal graphs as a directed acrylic graph structure provide more explicit and interpretable correlations between variables, thus enabling a better understanding of the underlying physical mechanisms and dynamic systems for time series~\cite{DBLP:journals/corr/abs-2310-20679, DBLP:conf/nips/KofinasNG21}.
As a widely applied causal structure in time series understanding and explanation, Granger Causality~\cite{granger1969investigating, DBLP:conf/kdd/ArnoldLA07} discovers causal relations among variables in an autoregressive (or time-lagged) manner. The discovered Granger causal structures can help many time series analysis tasks, like building parsimonious prediction models such as Earth System~\cite{runge2019detecting}. Moreover, real-world time series data can have many variables, and their causal relations can be even more complex, i.e., non-linear and instantaneous, which require complex causality discovery beyond the classic Granger model. Although some nascent non-linear (or neural) Granger models have been proposed~\cite{DBLP:journals/make/NautaBS19,DBLP:conf/cikm/XuHY19, DBLP:journals/pami/TankCFSF22, DBLP:conf/iclr/KhannaT20, DBLP:conf/cikm/HuangXYYWX20, DBLP:conf/aistats/PamfilSDPGBA20, DBLP:conf/iclr/MarcinkevicsV21, DBLP:journals/corr/abs-2202-02195}, how to effectively integrate instantaneous causal effects with neural Granger models has the great research potential~\cite{moneta2013causal, wild2010graphical, dahlhaus2003causality, DBLP:conf/kdd/MalinskyS18, DBLP:conf/uai/AssaadDG22} and remains largely underexplored~\cite{DBLP:conf/aistats/PamfilSDPGBA20, gong2023rhino}.

\begin{figure}[t]
\includegraphics[width=0.49\textwidth]{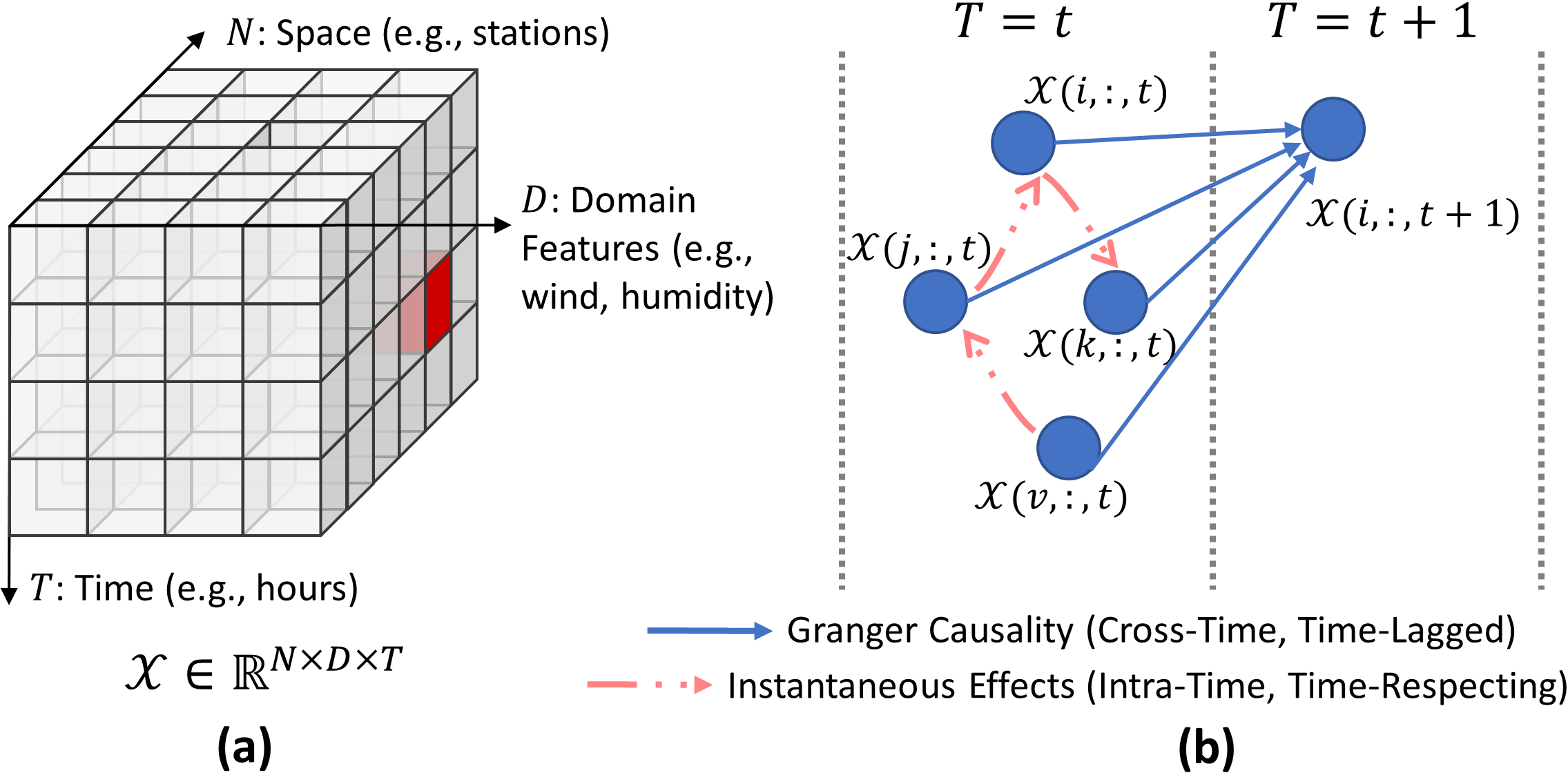}
\centering
\caption{(a) Tensor Time-Series Data: The Red Cell Means the Possible Anomaly. (b) Visualization of (Neural) Granger Causality's Time-Lagged Property without Instantaneous Effects.}
\vspace{-2mm}
\label{Fig:introduction}
\end{figure}

Motivated by the above analysis, in this paper, we start from the tensor time series data as shown in Figure~\ref{Fig:introduction}(a), in which the 3D structure contains higher dimensions than typical 2D multivariate time series data. For example, tensor time series can represent multivariate climate time series data (e.g., time plus temperature, wind, and atmospheric water content) with corresponding spatial information (e.g., longitude, latitude, and geocode). After that, we aim to build a comprehensive causality model for this tensor time series, which could not only capture non-linear and time-lagged causality (like the Granger model~\cite{granger1969investigating, DBLP:journals/pami/TankCFSF22}) but also offset the ignored instantaneous causal effects at each timestamp, as shown in Figure~\ref{Fig:introduction}(b). Our \textbf{ultimate goal} is to leverage the discovered comprehensive causality to understand the trend and latent distribution of the historical tensor time series and finally contribute to the analysis tasks like tensor time series forecasting and anomaly detection.

To this end, we first propose a comprehensive causal model named Time-Respecting Bayesian Network Augmented Neural Granger Causality, i.e., TBN Granger Causality.
Theoretically, discovering TBN Granger Causality relies on a bi-level optimization. The inner optimization discovers a sequence of Bayesian Networks at each timestamp $t$ respectively for representing the instantaneous causal effects among variables (i.e., which causality is responsible for the instantaneous feature generation). Then, the outer optimization realizes integrating time-respecting Bayesian Networks with time-lagged neural Granger causality in an autoregressive manner.

Second, to embed TBN Granger Causality into guiding the tensor time series analysis tasks like forecasting and anomaly detection, we propose an end-to-end deep generative model, called \textbf{\underline{T}}ime-\textbf{\underline{A}}ugmented \textbf{\underline{C}}ausal Time \textbf{\underline{S}}eries \textbf{\underline{A}}nalysi\textbf{\underline{S}} Model, i.e., \name.
\name\ fits more real-world application scenarios (e.g., climate or transportation) by investigating how to capture good causal structures {\em without} the ground-truth structures guidance.
Furthermore, \name\ is end-to-end, meaning that it can not only discover TBN Granger Causality from the observed time series but also seamlessly use the discovery to forecast future time series and detect possible anomalies.

To evaluate \name, we first use the synthetic benchmark, Lorenz-96~\cite{lorenz1996predictability}, to verify that \name\ can indeed discover ground-truth causal structures with high accuracy. Then, we extend to the real-world setting and test if \name\ can utilize the discovered causality to conduct tensor time series forecasting and identify anomalies.
We use four tensor time series datasets from the hourly climate benchmark database ERA5~\cite{hersbach2018era5} and align them with the extreme weather database of NOAA\footnote{\url{https://www.ncdc.noaa.gov/stormevents/ftp.jsp}} based on geoinformation and contribute a new benchmark for climate science. The results show that \name\ outperforms both state-of-the-art forecasting and detection baselines.


\section{Preliminary}
\textbf{Tensor Time Series}. As shown in Figure~\ref{Fig:introduction}(a), we have tensor time series data stored in $\mathcal{X} \in \mathbb{R}^{N \times D \times T }$. Note that a slice of $\mathcal{X}$, i.e., $\mathcal{X}(i, :, :) \in \mathbb{R}^{D \times T }$, $i \in \{1 \ldots, N\}$, is typically denoted as the common multivariate time series data~\cite{DBLP:conf/kdd/SuZNLSP19, DBLP:conf/icdm/ZhaoWDHCTXBTZ20}. In this way, tensor time series can be understood as multiple multivariate time series data.
Such tensor time series data can usually be found in the real world. For example, in each element $\mathcal{X}(i,d,t)$ of the nationwide weather data $\mathcal{X}$, $i \in \{1 \ldots, N\}$ can be the spatial locations (e.g., counties), $d \in \{1 \ldots, D\}$ can be the weather features (e.g., temperature and humidity), and $t \in \{1 \ldots, T\}$ can be the time dimension (e.g., hours). Throughout the paper, we use the calligraphic letter to denote a 3D tensor (e.g., $\mathcal{X}$) and the bold capital letter to denote a 2D matrix (e.g., $\bm{X}$). 

\textbf{Problem Definition}. In this paper, we aim to discover and utilize comprehensive causal structures for tensor time-series analysis tasks, including forecasting and anomaly detection.
To be more specific, given the tabular data $\mathcal{X} \in \mathbb{R}^{N \times D \times T}$ as shown in Figure~\ref{Fig:introduction}, we aim to forecast the future data $\mathcal{X}' \in \mathbb{R}^{N \times D \times \tau}$, where $\tau$ is a forecasting window. Additionally, with the forecasted $\mathcal{X}'$, we also aim to detect if $\mathcal{X}'$ contains abnormal values. 

\section{\name: Discovering TBN Granger Causality via Generative Learning}
\begin{figure*}[h]
\includegraphics[width=0.94\textwidth]{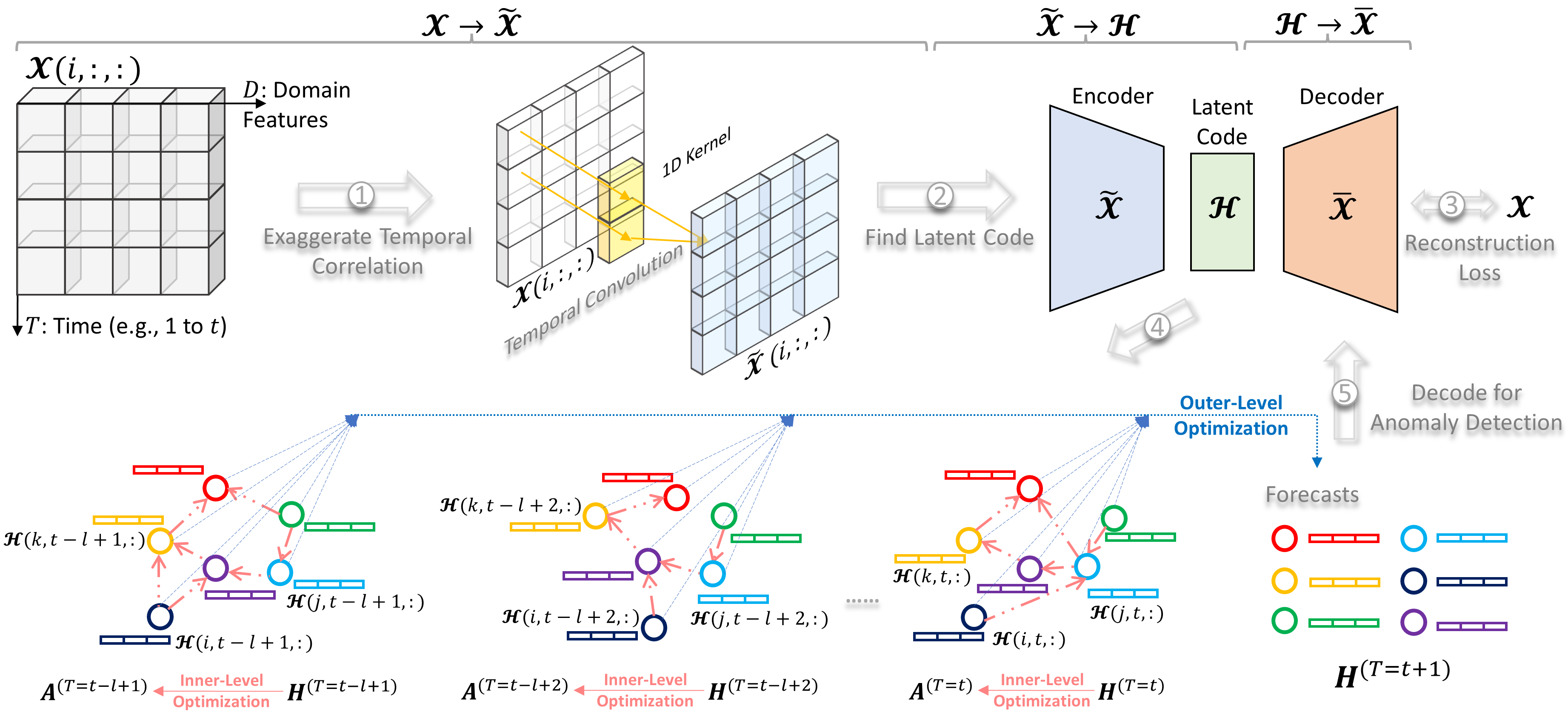}
\centering
\caption{Working Flow of \name: Discovering and Utilizing the TBN Granger Causality through a Bi-Level Optimization for Tensor Time Series Forecasting and Anomaly Detection.}
\vspace{-3mm}
\label{Fig:framework}
\end{figure*}

In this section, we introduce how \name\ discovers TBN Granger Causality in the historical tensor time series and utilizes it to guide tensor time series forecasting and anomaly detection. The overall framework of \name\ is shown in Figure~\ref{Fig:framework}.

The upper component of Figure~\ref{Fig:framework} represents the data preprocessing part (i.e., converting raw input $\mathcal{X}$ to latent representation $\mathcal{H}$) of \name\ through a pre-trained autoencoder. The goal of this component is leveraging comprehensive causality (e.g., TBN Granger Causality) to achieve seamless forecasting and anomaly detection. The theoretical reasoning and necessity are introduced in Sec.3.3, and the empirical validation is demonstrated in Appendix B.2. 

The lower component of Figure~\ref{Fig:framework} shows how \name\ discovers TBN Granger Causality in the historical tensor time series (in the form of $\mathcal{H}$ other than $\mathcal{X}$) and generates future tensor time series. In brief, the optimization of \name\ is bi-level. First, the inner optimization captures instantaneous effects among variables at each timestamp, respectively, which describes the inner-time feature generation. These causal structures are then stored in the form of a sequence of Bayesian Networks. The details are introduced in Sec.3.1. Second, the outer optimization discovers the Neural Granger Causality among variables in a time window with the support of a sequence of Bayesian Networks (i.e., TBN Granger Causality). After introducing details in Sec.3.2, we derive the formal equation of TBN Granger Causality, Eq.~\ref{eq:forecast}.

\subsection{Inner Optimization of \name\ for Identifying Instantaneous Causal Relations in Time Series}
Generally speaking, the inner optimization produces a sequence of Bayesian Networks for each observed timestamp. At time $t$, the instantaneous causality is discovered based on input features $\mathcal{H}(:,:,t) = \bm{H}^{(t)} \in \mathbb{R}^{N \times H}$, and is represented by a directed acyclic graph $\mathcal{G}^{(t)} = (\bm{A}^{(t)} \in \mathbb{R}^{N \times N}, \bm{H}^{(t)}\in \mathbb{R}^{N \times H})$. To be specific, $\bm{A}^{(t)}$ is a weighted adjacency matrix of the Bayesian Network at time $t$, and each cell represents the coefficient of causal effects between variables $u$ and $v \in \{1, \ldots, N\}$. The features (e.g., $\mathcal{H}(v,:,t)$) are transformed from the input raw features (e.g., $\mathcal{X}(v,:,t)$). The transformation is causality-agnostic but necessary for downstream time series analysis tasks, with details introduced in Sec.3.3.

The reasoning for discovering the instantaneous causal effects in the form of the Bayesian Network originates from a widely adopted assumption of causal graph learning~\cite{DBLP:conf/nips/ZhengARX18, DBLP:conf/icml/YuCGY19, DBLP:journals/csur/GuoCLH020, DBLP:journals/corr/abs-2202-02195, gong2023rhino}: there exists a ground-truth causal graph $\mathbf{S}^{(t)}$ that specifies instantaneous parents of variables to recover their value generating process. Therefore, in our inner optimization, the goal is to discover the causal structure $\mathbf{S}^{(t)}$ at each time $t$ by recovering the generation of input features $\bm{H}^{(t)}$. Specifically, given the observed $\bm{H}^{(t)}$, we aim to estimate a structure $\bm{A}^{(t)}$, through which a certain distribution $\bm{Z}^{(t)}$ could generate $\bm{H}^{(t)}$ for $t \in \{1, \ldots, T\}$. In this way, the instantaneous causal effects are discovered, and the corresponding structures are encoded in $\bm{A}^{(t)}$. The generation function is expressed as follows.
\begin{equation}
    \sum_{t} log \mathcal{P}(\bm{H}^{(t)}) = \sum_{t}log \int \mathcal{P}(\bm{H}^{(t)} | \bm{Z}^{(t)}) \mathcal{P}(\bm{Z}^{(t)}) d\bm{Z}^{(t)}
\label{eq: generative_model}
\end{equation}
where the generation likelihood $\mathcal{P}(\bm{H}^{(t)} | \bm{Z}^{(t)})$ also takes $\bm{A}^{(t)}$ as input. The complete formula is shown in Eq.~\ref{eq: DAG_Decoder}.

For Eq~\ref{eq: generative_model}, on the one hand, it is hard to get the prior distribution $\mathcal{P}(\bm{Z}^{(t)})$, which is highly related to the distribution of ground-truth causal graph distribution $\mathcal{P}(\mathbf{S}^{(t)})$ at time $t$~\cite{DBLP:journals/corr/abs-2202-02195}. On the other hand, for the generation likelihood $\mathcal{P}(\bm{H}^{(t)} | \bm{Z}^{(t)})$, the actual posterior $\mathcal{P}(\bm{Z}^{(t)} | \bm{H}^{(t)})$ is also intractable. Thus, we resort to the variational autoencoder (VAE)~\cite{DBLP:journals/corr/KingmaW13}. In this way, the actual posterior $\mathcal{P}(\bm{Z}^{(t)} | \bm{H}^{(t)})$ can be replaced by the variational posterior $\mathcal{Q}(\bm{Z}^{(t)} | \bm{H}^{(t)})$, and the prior distribution $\mathcal{P}(\bm{Z}^{(t)})$ is approximated by a Gaussian distribution. Furthermore, the inside encoder and decoder modules should take the structure $\bm{A}^{(t)}$ as the input. This design can be realized by various off-the-shelf variational graph autoencoders such as VGAE~\cite{DBLP:journals/corr/KipfW16a}, etc. However, the inner optimization is coupled with the outer optimization, i.e., the instantaneous causality will be integrated with cross-time Granger causality to make inferences. The inner complex neural architectures and parameters may render the outer optimization module hard to train, especially when the outer module itself needs to be complex. Therefore, we extend the widely-adopted linear Structural Equation Model (SEM)~\cite{DBLP:conf/nips/ZhengARX18, DBLP:conf/icml/YuCGY19, DBLP:journals/corr/abs-2202-02195, gong2023rhino} to the time-respecting setting as follows.

For $\mathcal{Q}(\bm{Z}^{(t)} | \bm{H}^{(t)})$, the encoder equation is expressed as
\begin{equation}
    \bm{Z}^{(t)} = (\bm{I} - {\bm{A}^{(t)}}^{\top})f_{\theta_{enc}^{(t)}}(\bm{H}^{(t)})
\label{eq: DAG_Encoder}
\end{equation}
For $\mathcal{P}(\bm{H}^{(t)} | \bm{Z}^{(t)})$, the decoder equation is expressed as
\begin{equation}
    \bm{H}^{(t)} = f_{\theta_{dec}^{(t)}}((\bm{I} - {\bm{A}^{(t)}}^{\top})^{-1}\bm{Z}^{(t)})
\label{eq: DAG_Decoder}
\end{equation}
As analyzed above\footnote{The complete forms of $\mathcal{Q}(\bm{Z}^{(t)} | \bm{H}^{(t)})$ and $\mathcal{P}(\bm{H}^{(t)} | \bm{Z}^{(t)})$ are $\mathcal{Q}_{A^{(t)}}(\bm{Z}^{(t)} | \bm{H}^{(t)})$ and $\mathcal{P}_{A^{(t)}}(\bm{H}^{(t)} | \bm{Z}^{(t)})$, we omit the subscript ${A^{(t)}}$ for brevity.}, $f_{\theta_{enc}^{(t)}}$ and $f_{\theta_{dec}^{(t)}}$ do not need complicated neural architectures. Therefore, we can use two-layer MLPs for them.
Then, the objective function $\mathcal{L}_{DAG}^{(t)}$ for discovering the instantaneous causality at time $t$ is expressed as follows, which corresponds to the inner optimization.
\begin{equation}
\begin{split}
    \min\limits_{\theta^{(t)}_{enc},\theta^{(t)}_{dec}, \bm{A}^{(t)}} \mathcal{L}_{DAG}^{(t)} = ~D_{KL}(\mathcal{Q}(\bm{Z}^{(t)} | \bm{H}^{(t)}) \|& \mathcal{P}(\bm{Z}^{(t)}))\\ 
    - \mathbb{E}_{\mathcal{Q}(\bm{Z}^{(t)} | \bm{H}^{(t)})}[\text{log} \mathcal{P}(\bm{H}^{(t)} &| \bm{Z}^{(t)})] \\
     \text{s.t.}~ \sum_{t}\text{Tr}[(\bm{I} + \bm{A}^{(t)} \circ \bm{A}^{(t)})^{N}] - N =0, ~\text{for}~ t & \in \{1, \ldots, T\}
\end{split}
\label{eq: inner_opt}
\end{equation}
where the first term in $\mathcal{L}_{DAG}^{(t)}$ is the KL-divergence measuring the distance between the distribution of generated $\bm{Z}^{(t)}$ and the pre-defined Gaussian, and the second term is the reconstruction loss between the generated $\bm{Z}^{(t)}$ with the original input $\bm{H}^{(t)}$. Note that there is an important constraint, i.e.,  $\text{Tr}[(\bm{I} + \bm{A}^{(t)} \circ \bm{A}^{(t)})^{N}] - N =0$, on $\bm{A}^{(t)} \in \mathbb{R}^{N \times N}$. $\text{Tr} (\cdot)$ is the trace of a matrix, and $\circ$ denotes the Hadamard product. The meaning of the constraint is explained as follows.
The constraint in Eq.~\ref{eq: inner_opt}, i.e., $\text{Tr}[(\bm{I} + \bm{A}^{(t)} \circ \bm{A}^{(t)})^{N}] - N =0$ regularizes the acyclicity of $\bm{A}^{(t)}$ during the optimization process, i.e., the learned $\bm{A}^{(t)}$ should not have any possible closed-loops at any length.
\begin{lemma}
Let $\bm{A}^{(t)}$ be a weighted adjacency matrix (negative weights allowed). $\bm{A}^{(t)}$ has no $N$-length loops, if $\text{Tr}[(\bm{I} + \bm{A}^{(t)} \circ \bm{A}^{(t)})^{N}] - N =0$.
\label{eq:acyclicity}
\end{lemma}

The intuition is that there will be no $k$-length path from node $u$ to node $v$ on a binary adjacency matrix $\}(u,v) = 0$. Compared with original acyclicity constraints in ~\cite{DBLP:conf/icml/YuCGY19}, our Lemma~\ref{eq:acyclicity} gets rid of the $\lambda$ condition. Then we can denote $\alpha({A}^{(t)}) = \text{Tr}[(\bm{I} + \bm{A}^{(t)} \circ \bm{A}^{(t)})^{N}] - N$ and use Lagrangian optimization for Eq.~\ref{eq: inner_opt} as follows.
\begin{equation}
\begin{split}
    \min\limits_{\theta^{(t)}_{enc},\theta^{(t)}_{dec}, \bm{A}^{(t)}} \mathcal{L}_{DAG}^{(t)} =  ~D_{KL}(\mathcal{Q}(\bm{Z}^{(t)} | \bm{H}^{(t)}) \| \mathcal{P}(\bm{Z}^{(t)})) \\
    - \mathbb{E}_{\mathcal{Q}(\bm{Z}^{(t)} | \bm{H}^{(t)})}[\text{log} \mathcal{P}(\bm{H}^{(t)} | \bm{Z}^{(t)})] \\
    + \lambda ~\alpha({A}^{(t)}) + \frac{c}{2}|\alpha({A}^{(t)})|^{2},~ ~\text{for}~ t \in \{1, \ldots, T\}
\end{split}
\label{eq: complete_inner_opt}
\end{equation}
where $\lambda$ and $c$ are two hyperparameters, and larger $\lambda$ and $c$ enforce $\alpha(\bm{A}^{(t)})$ to be smaller.

\begin{theorem}
    If the ground-truth instantaneous causal graph $\mathbf{S}^{(t)}$ at time $t$ generates the features of variables following the normal distribution, then the inner optimization (i.e., Eq.~\ref{eq: inner_opt}) can identify $\mathbf{S}^{(t)}$ under the standard causal discovery assumptions~\cite{DBLP:journals/corr/abs-2202-02195}.
    \label{eq: discovery}
\end{theorem}

\subsection{Outer Optimization of \name\ for Integrating Instantaneous Causality with Neural Granger}
Given the inner optimization, Bayesian Networks can be obtained at each timestamp $t$, which means that multiple instantaneous causalities are discovered. Thus, in the outer optimization, we integrate these evolving Bayesian Networks into Granger Causality discovery.
First, the classic Granger Causality~\cite{granger1969investigating} is discovered in the form of the variable-wise coefficients across different timestamps (i.e., a time window) through the autoregressive prediction process. The prediction based on the linear Granger Causality~\cite{granger1969investigating} is expressed as follows.
\begin{equation}
    \bm{H}^{(t)} = \sum_{l=1}^{L} \bm{W}^{(l)}\bm{H}^{(t-l)} + \bm{e}^{(t)}
\label{eq:graner_causality}
\end{equation}
where $\bm{H}^{(t)} \in \mathbb{R}^{N \times D}$ denotes the features of $N$ variables at time $t$, $\bm{e}^{(t)}$ is the noise, and $L$ is the pre-defined time lag indicating how many past timestamps can affect the values of $\bm{H}^{(t)}$. Weight matrix $\bm{W}^{(l)} \in \mathbb{R}^{N \times N}$ stores the cross-time coefficients captured by Granger Causality, i.e., matrix $\bm{W}^{(l)}$ aligns the variables at time $t-l$ with the variables at time $t$. To compute those weights, several linear methods are proposed, e.g., vector autoregressive model~\cite{DBLP:conf/kdd/ArnoldLA07}.

Facing non-linear causal relationships, neural Granger Causality discovery~\cite{DBLP:journals/pami/TankCFSF22} is recently proposed to explore the nonlinear Granger Causality effects. The general principle is to represent causal weights $\bm{W}$ by deep neural networks. To integrate instantaneous effects with neural Granger Causality discovery, our TBN Granger Causality is expressed as follows.
\begin{equation}
\label{eq:forecast}
    \hat{H}(i,:)^{(t)} =  f_{\Theta_{i}}[(\bm{A}^{(t-1)}, \bm{H}^{(t-1)}), \ldots, (\bm{A}^{(t-L)}, \bm{H}^{(t-L)})]
\end{equation}
where $L$ is the lag (or window size) in the Granger Causality, and $i$ is the index of the $i$-th variable. $f_{\Theta_{i}}$ is a neural computation unit with all parameters denoted as $\Theta_{i}$, whose input is an $L$-length time-ordered sequence of $(\bm{A}, \bm{H})$. And $f_{\Theta_{i}}$ is responsible for discovering the TBN Granger Causality for variable $i$ at time $t$ from all variables that occurred in the past time lag $l$. The choice of neural unit $f_{\Theta_{i}}$ is flexible, such as MLP and LSTM~\cite{DBLP:journals/pami/TankCFSF22}. Different neural unit choices correspond to different causality interpretations. In our proposed \name\ model, we use graph recurrent neural networks~\cite{DBLP:journals/tnn/WuPCLZY21}, and the causality interpretations are introduced in Sec 3.3.

In the outer optimization, to evaluate the prediction under the TBN Granger Causality, we use the mean absolute error (MAE) loss on the prediction and the ground truth, which is effective and widely applied to time-series forecasting tasks~\cite{DBLP:conf/iclr/LiYS018, DBLP:conf/iclr/Shang0B21}.
\begin{equation}
   \min_{\Theta_{i}, \bm{A}^{(t-1)}, \ldots, \bm{A}^{(t-l)} } \mathcal{L}_{pred} = \sum_{i} \sum_{t} | H(i,:)^{(t)} - \hat{H}(i,:)^{(t)}|
\label{eq: outer_opt}
\end{equation}
where $\Theta_{i}, \bm{A}^{(t-1)}, \ldots, \bm{A}^{(t-l)}$ are all the parameters for the prediction $\hat{H}(i,:)^{(t)}$ of variable $i$ at time $t$. The composition and update rules are expressed below.

\textbf{For updating $f_{\Theta_i}$}, we employ the recurrent neural structure to fit the input sequence. Moreover, the sequential inputs also contain the structured data $\bm{A}$. Therefore, we use the graph recurrent neural architecture~\cite{DBLP:conf/iclr/LiYS018} because it is designed for directed graphs, whose core is a gated recurrent unit~\cite{DBLP:journals/corr/ChungGCB14}.
\begin{equation}
\begin{split}
    & \bm{R}^{(t)} = \text{sigmoid} (\bm{W}_{\bm{R} \ast \bm{A}^{(t)}} [\bm{H}^{(t)} \oplus \bm{S}^{(t-1)}] + \bm{b}_{R})\\
    & \bm{C}^{(t)} = \text{tanh} (\bm{W}_{\bm{C} \ast \bm{A}^{(t)}} [\bm{H}^{(t)} \oplus (\bm{R}^{(t)} \odot \bm{S}^{(t-1)})] + \bm{b}_{C})\\
    & \bm{U}^{(t)} = \text{sigmoid} (\bm{W}_{\bm{U} \ast \bm{A}^{(t)}} [\bm{H}^{(t)} \oplus \bm{S}^{(t-1)}] + \bm{b}_{U})\\
    & \bm{S}^{(t)} = \bm{U}^{(t)} \odot \bm{S}^{(t-1)} + (\bm{I} - \bm{U}^{(t)}) \odot \bm{C}^{(t)}
\end{split}
\label{eq: gates}
\end{equation}
where $\bm{R}^{(t)}$, $\bm{C}^{(t)}$, and $\bm{U}^{(t)}$ are three parameterized gates, with corresponding weights $\bm{W}$ and bias $\bm{b}$. $\bm{H}^{(t)}$ is the input, and $\bm{S}^{(t)}$ is the hidden state. Gates $\bm{R}^{(t)}$, $\bm{C}^{(t)}$, and $\bm{U}^{(t)}$ share the similar structures. For example, in $\bm{R}^{(t)}$, the graph convolution operation for computing the weight $\bm{W}_{\bm{R} \ast \bm{A}^{(t)}}$ is defined as follows, and the same computation applies to gates $\bm{U}^{(t)}$ and $\bm{C}^{(t)}$.
\begin{equation}
    \bm{W}_{\bm{R} \ast \bm{A}^{(t)}} = \sum_{k=0}^{K} ~\theta_{k,1}^{R}({\bm{D}^{(t)}_{out}}^{-1}\bm{A}^{(t)})^{k} +\theta_{k,2}^{R}({\bm{D}^{(t)}_{in}}^{-1}{\bm{A}^{(t)}}^{\top})^{k}
\end{equation}
where $\theta_{k,1}^{R}$, $\theta_{k,2}^{R}$ are learnable weight parameters; scalar $k$ is the order for the stochastic diffusion operation (i.e., similar to steps of random walks); ${\bm{D}^{(t)}_{out}}^{-1}\bm{A}^{(t)}$ and ${\bm{D}^{(t)}_{in}}^{-1}{\bm{A}^{(t)}}^{\top}$ serve as the transition matrices with the in-degree matrix $\bm{D}^{(t)}_{in}$ and the out-degree matrix $\bm{D}^{(t)}_{out}$; $-1$ and $\top$ are inverse and transpose operations.

\textbf{For updating each of $\{\bm{A}^{(t-1)}, \ldots, \bm{A}^{(t-l)}\}$}, we take $\bm{A}^{(t-l)}$ as an example to illustrate. The optimal $\bm{A}^{(t-l)}$ stays in the space of $\{0,1\}^{N \times N}$. To be specific, each edge $A^{(t-l)}(i,j)$ can be parameterized as $\theta_{i,j}^{(t-l)}$ following the Bernoulli distribution. However, $N^{2}l$ is hard to scale, and the discrete variables are not differentiable. Therefore, we adopt the Gumbel reparameterization from~\cite{DBLP:conf/iclr/JangGP17, DBLP:conf/iclr/MaddisonMT17}. It provides a continuous approximation for the discrete distribution, which has been widely used in the graph structure learning~\cite{DBLP:conf/icml/KipfFWWZ18, DBLP:conf/iclr/Shang0B21}. The general reparameterization form can be written as $A^{(t-l)}(i,j) = softmax (FC((H(i,:)^{(t-l)}||H(j,:)^{(t-l)}) + g)/\xi)$, where $FC$ is a feedforward neural network, $g$ is a scalar drawn from a Gumbel$(0,1)$ distribution, and $\xi$ is a scaling hyperparameter. Different from~\cite{DBLP:conf/icml/KipfFWWZ18, DBLP:conf/iclr/Shang0B21}, in our setting, the initial structure input is constrained by the causality discovery, which originates from the inner optimization step. Hence, the structure learning in the outer optimization takes the adjacency matrix from the inner optimization as the initial input, which is
\begin{equation}
    A^{(t-l)}_{outer}(i,j) = softmax (A^{(t-l)}_{inner}(i,j) + \bm{g})/\xi)
\end{equation}
where $A^{(t)}_{inner}(i,j)$ is the structure learned by our inner optimization through Eq.~\ref{eq: inner_opt}, $A^{(t)}_{outer}(i,j)$ is the updated structure, and $\bm{g}$ is a vector of i.i.d samples drawn from a Gumbel$(0,1)$ distribution. In outer optimization, Eq.~\ref{eq: outer_opt} fine-tunes the evolving Bayesian Networks to make the intra-time causality fit the cross-time causality well. Note that, the outer optimization w.r.t. $\bm{A}^{(t)}$ may break the acyclicity, and another round of inner optimization may be necessary.

\subsection{Deployment of \name\ for Time Series Forecasting and Anomaly Detection}


\vspace{-2mm}
In this section, we introduce how \name\ achieves tensor time series forecasting and anomaly detection in threefold: data preprocessing, neural architecture selection, and training procedure.

\textbf{First (data preprocessing)}, in addition to forecasting, \name\ is also for anomaly detection. Thus, we design the hidden feature $\mathcal{H}$ extraction in \name\ motivated by the Extreme Value Theory~\cite{beirlant2004statistics} or so-called Extreme Value Distribution in stream~\cite{ DBLP:conf/kdd/SifferFTL17}.

\begin{remark}
According to the Extreme Value Distribution~\cite{fisher1928limiting}, under the limiting forms of frequency distributions, extreme values have the same kind of distribution, regardless of original distributions.
\label{remark}
\end{remark}
\vspace{-2mm}

An example~\cite{DBLP:conf/kdd/SifferFTL17} can help interpret and understand the Extreme Value Distribution theory. Maximum temperatures or tide heights have more or less the same distribution even though the distributions of temperatures and tide heights are not likely to be the same. As rare events have a lower probability, there are only a few possible shapes for a general distribution to fit.
Inspired by this observation, we can design a simple but effective module in \name\ to achieve anomaly detection, i.e., a pre-trained autoencoder model that tries to explore the distribution of normal features in $\mathcal{X}$ as shown in Figure~\ref{Fig:framework}. As long as this autoencoder model can capture the latent distribution for normal events, then the generation probability of a piece of time series data can be utilized as the condition for detecting anomaly patterns. This is because the extreme values are identified with a remarkably low generation probability. To be specific, after the forecast $\bm{H}^{(t)}$ is output, the generation probability of $\bm{H}^{(t)}$ into $\bm{X}^{(t)}$ through the pre-trained autoencoder can be used to detect the anomalies at $t$.

\textbf{Second (neural architecture selection)}, we encode $f_{\Theta_{i}}$ into a sequence-to-sequence model~\cite{DBLP:conf/nips/SutskeverVL14}. That is, given a time window (or time lag), \name\ could forecast the corresponding features for the next time window.
Moreover, with $\bm{W}^{(l)}$ in Eq.~\ref{eq:graner_causality} and $f_{\Theta_{i}}$ in Eq.~\ref{eq:forecast}, we can observe that the classical linear Granger Causality $\bm{W}^{(l)}$ can be discovered for each time lag. In other words, each time lag has its own discovered coefficients, but $f_{\Theta_{i}}$ is shared by all time lags. This sharing manner is designed for scalability and is called Summary Causal Graph~\cite{DBLP:conf/iclr/MarcinkevicsV21, DBLP:conf/uai/AssaadDG22}. The underlying intuition is that the causal effects mainly depend on the near timestamps. Further, for the neural Granger Causality interpretation in $f_{\Theta_{i}}$, we follow the rule~\cite{DBLP:journals/pami/TankCFSF22} that if the $j$-th row of ($\bm{W}_{\bm{R} \ast \bm{A}^{(t)}}$, $\bm{W}_{\bm{C} \ast \bm{A}^{(t)}}$, and $\bm{W}_{\bm{U} \ast \bm{A}^{(t)}}$) are zeros, then variable $j$ is not the Granger-cause for variable $i$ in this time window.

\textbf{Third (training procedure)}, as shown in Figure~\ref{Fig:framework}, the autoencoder can be pre-trained with reconstruction loss (e.g., MSE) ahead of the inner and outer optimization, to obtain $\mathcal{H}$ for the feature latent distribution representation. By utilizing all input $\mathcal{H}$, the inner optimization learns the sequential Bayesian Networks, and the outer optimization aligns Bayesian Networks with the neural Granger Causality to produce all the forecast $\mathcal{H}'$. The inner and outer optimization can be trained interchangeably.

\section{Experiments}
The ground-truth causality discovery experiments in the synthetic benchmark, Lorenz 96 System~\cite{lorenz1996predictability}, are shown in Appendix~\ref{sec:empirical_analysis}.1, where our \name\ can capture the true causality with the competitive high accuracy.
Then, in this section, we test \name\ on utilizing its discovery for time series forecasting and anomaly detection.

\begin{table*}[t]
\caption{Forecasting Error (MAE, $10^{-2}$)}
\vspace{-1mm}
\centering
\scalebox{0.95}{
\begin{tabular}{ccccc}
\hline
      & ERA5-2017 $(\downarrow)$          & ERA5-2018 $(\downarrow)$          & ERA5-2019 $(\downarrow)$          & ERA5-2020 $(\downarrow)$ \\ \hline
GRU   & 1.8834 $\pm$ 0.0126               & 1.9764 $\pm$ 0.1466               & 1.6194 $\pm$ 0.2645               & 1.7859 $\pm$ 0.2324           \\
DCRNN & 0.0819 $\pm$ 0.0025               & 0.0797 $\pm$ 0.0049               & 0.0799 $\pm$ 0.0035               & 0.0826 $\pm$ 0.0033           \\
GTS   & 0.0777 $\pm$ 0.0054               & 0.0766 $\pm$ 0.0029               & 0.0760 $\pm$ 0.0031               & 0.0742 $\pm$ 0.0021           \\
\name\ & 0.0496 $\pm$ 0.0017              & 0.0499 $\pm$ 0.0017               & 0.0502 $\pm$ 0.0016               & 0.0488 $\pm$ 0.0019 \\
ST-SSL & 0.0345 $\pm$ 0.0051               & 0.0330 $\pm$ 0.0018               & 0.0361 $\pm$ 0.0021               & 0.0348 $\pm$ 0.0020           \\
\name++ & $\textbf{0.0271}$ $\pm$ $\textbf{0.0004}$ & $\textbf{0.0276}$ $\pm$ $\textbf{0.0004}$ & $\textbf{0.0282}$ $\pm$ $\textbf{0.0003}$ & $\textbf{0.0265}$ $\pm$ $\textbf{0.0004}$ \\\hline
\end{tabular}}
\label{tb:forecasting}
\end{table*}

\subsection{Experiment Setup}

\textbf{Datasets}. Our forecasting data (i.e., hourly tensor time series data) originates from climate domain benchmark ERA5~\cite{hersbach2018era5}\footnote{\url{https://cds.climate.copernicus.eu/cdsapp\#!/home}}. To be specific, we select four datasets covering 45 weather features (i.e., wind gusts, rain, etc.) from 238\footnote{100 of 238 counties are top-ranked counties for the thunderstorm (anomaly label) frequency, and the rest are randomly selected.} counties in the United States of America during 2017--2020. Moreover, we choose thunderstorms as the anomaly pattern to be detected after forecasting. The thunderstorm record is identified in NOAA database\footnote{\url{https://www.ncdc.noaa.gov/stormevents/ftp.jsp}} hourly and nationwide, i.e., 1 means a thunderstorm happens in the corresponding hour at a certain location, and 0 means no thunderstorm happens. 
We processed the geocode to align weather features in ERA5 with anomaly patterns in NOAA. The geographic distribution and anomaly pattern frequency distribution are shown in Appendix~\ref{sec:dataset}.

\textbf{Baselines}.
Besides the causality discovery baseline in Appendix~\ref{sec:empirical_analysis}, the \textbf{first} category is for tensor time series forecasting: (1) GRU~\cite{DBLP:journals/corr/ChungGCB14} is a classical sequence to sequence generation model. (2) DCRNN~\cite{DBLP:conf/iclr/LiYS018} is a graph convolutional recurrent neural network, of which the input graph structure is given, not causal, and static (i.e., shared by all timestamps). In this viewpoint, we let each node randomly distribute its unit weights to others. (3) GTS~\cite{DBLP:conf/iclr/Shang0B21} is also a graph convolutional recurrent neural network that does not need the input graph but learns the structure based on the node features, but the learned structure is also shared by all timestamps and is not causal. To compare the performance of DCGNN~\cite{DBLP:conf/iclr/LiYS018} and GTS~\cite{DBLP:conf/iclr/Shang0B21} with \name, causality is the control variable since we make all the rest (e.g., neural network type, number of layers, etc.) identical for them.
The \textbf{second} category is for anomaly detection on tensor time series: (1) DeepSAD~\cite{DBLP:conf/iclr/RuffVGBMMK20}, (2) DeepSVDD~\cite{DBLP:conf/icml/RuffGDSVBMK18}, and (3) DROCC~\cite{DBLP:conf/icml/GoyalRJS020}. Since these three have no forecast abilities, we let them use the ground-truth observations, and our \name\ utilizes the forecast features during anomaly detection experiments. Also, these three baselines are designed for multi-variate time-series data, not tensor time-series. Thus, we flatten our tensor time series along the spatial dimension and report the average performance for these three baselines over all locations.

Next, we introduce forecasting and anomaly detection performance. Details about split and hyperparameters are in Appendix~\ref{sec:implementation}. More ablation studies can be found in Appendix~\ref{sec:empirical_analysis}.3.

\vspace{-1mm}

\subsection{Forecasting Performance}
In Table~\ref{tb:forecasting}, we present the forecasting performance in terms of mean absolute error (MAE) on the testing data of three algorithms, namely DCGNN~\cite{DBLP:conf/iclr/LiYS018}, GTS~\cite{DBLP:conf/iclr/Shang0B21}, ST-SSL~\cite{DBLP:conf/aaai/JiW00XWZZ23}, our \name, and \name++ (i.e., \name\ with persistence forecast constraints). Here, we set the time window as 24, meaning that we use the past 24 hours tensor time series to forecast the future 24 hours in an autoregressive manner.
Moreover, for baselines and \name, we set $f_{\Theta_{i}}$ in Eq.\ref{eq:forecast} shared by all weather variables to ensure the scalability, such that we do not need to train $N$ recurrent graph neural networks for a single prediction. In Table~\ref{tb:forecasting}, we can observe a general pattern that our \name\ outperforms the baselines with GTS performing better than DCGNN. For example, with 2017 as the testing data, our \name\ performs 39.44\% and 36.16\% better than DCRNN and GTS.
An explanation is that the temporally fine-grained causal relationships can contribute more to the forecasting accuracy than non-causal directed graphs, since DCGNN, GTS, and our \name\ all share the graph recurrent manner. \name\, however, discovers causalities at different timestamps, while DCGNN and GTS use feature similarity based connections. Moreover, ST-SSL achieves competitive forecasting performance via contrastive learning on time series. Motivated by contrastive manner, \name++ is proposed by persistence forecast constraints. That is, the current forecast of \name\ is further calibrated by its nearest time window (i.e., the last 24 hours in our setting). The detailed implementation is provided in Appendix~\ref{sec:implementation}.

\vspace{-3mm}

\begin{table*}[t]
\caption{Anomaly Detection Performance (AUC-ROC)}
\vspace{-2mm}
\centering
\scalebox{0.95}{
\begin{tabular}{ccccc}
\hline
         & NOAA-2017 $(\uparrow)$                    & NOAA-2018 $(\uparrow)$                    & NOAA-2019 $(\uparrow)$                    & NOAA-2020 $(\uparrow)$ \\ \hline
DeepSAD  & 0.5305 $\pm$ 0.0481                       & 0.5267 $\pm$ 0.0406                       & 0.5563 $\pm$ 0.0460                       & 0.6420 $\pm$ 0.0054           \\
DeepSVDD & 0.5201 $\pm$ 0.0045                       & 0.5603 $\pm$ 0.0111                       & $\textbf{0.6784}$ $\pm$ $\textbf{0.0112}$ & 0.5820 $\pm$ 0.0205           \\
DROCC    & 0.5319 $\pm$ 0.0661                       & 0.5103 $\pm$ 0.0147                       & 0.6236 $\pm$ 0.0992                       & 0.5630 $\pm$ 0.1082           \\
\name\   & $\textbf{0.5556}$ $\pm$ $\textbf{0.0010}$ & $\textbf{0.5685}$ $\pm$ $\textbf{0.0011}$ & 0.6298 $\pm$ 0.0184                       & $\textbf{0.6745}$ $\pm$ $\textbf{0.0185}$ \\ \hline
\end{tabular}}
\label{tb:anomaly detection}
\vspace{-2mm}
\end{table*}

\begin{figure}[h]
\includegraphics[width=0.5\textwidth]{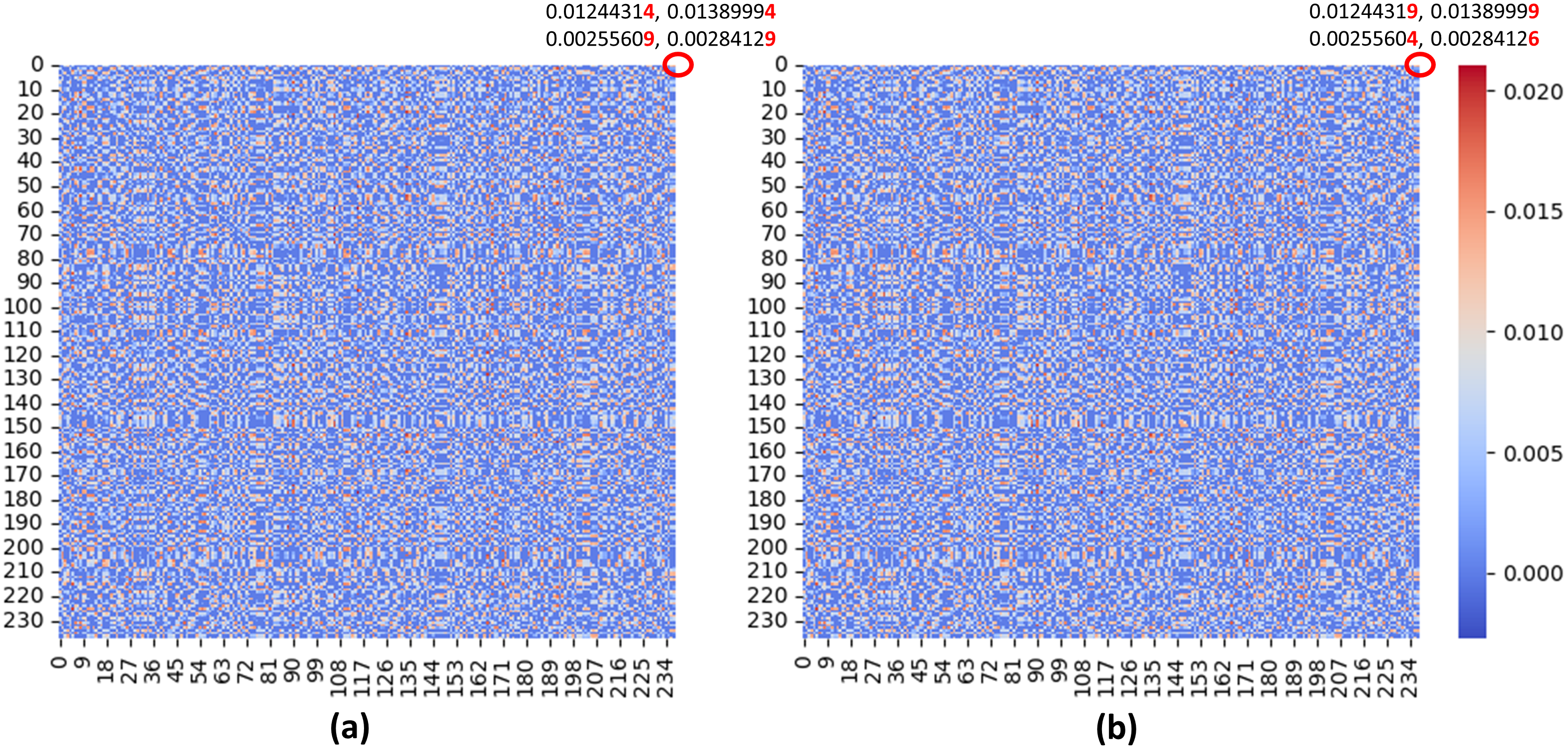}
\centering
\caption{Time-Respecting Bayesian Networks of at the Same Hour of Two Consecutive Days.}
\vspace{-3mm}
\label{Fig:dag_vis}
\end{figure}

To evaluate our explanation, we visualize causal connections at different times in Figure~\ref{Fig:dag_vis}. Specifically, we show the Bayesian Network of 238 counties at the same hour on two consecutive days in the training data (i.e., May 1st and May 2nd, 2018). Interestingly, we can observe that two patterns in Figure~\ref{Fig:dag_vis} are almost identical at first glance. That could be the reason why DCRNN and GTS can perform well using the static structure. However, upon closer inspection, we find that these two are quite different to some extent if we zoom in, such as, in the upper right corner. Although the values have a tiny divergence, their volume is quite large. In two matrices of Figure~\ref{Fig:dag_vis}, the number of different cells is 28,509, and the corresponding percentage is $\frac{28509}{238 \times 238} \approx 0.5033$. We suppose that discovering those value-tiny but volume-big differences makes \name\ outperform, to a large extent.


\subsection{Anomaly Detection}
After forecasting, we can have the hourly forecast of weather features at certain locations, denoted as $\mathcal{X}'$. Then, we use the encoder-decoder model in Figure~\ref{Fig:framework} to calculate the feature-wise generation probability using mean squared error (MSE) between $\mathcal{X}'$ and its generation $\bar{\mathcal{X}}'$. Thus, we can calculate the average of feature-wise generation probability as the condition of anomalies to identify if an anomaly weather pattern (e.g., a thunderstorm) happens in an hour in a particular location. In Table~\ref{tb:anomaly detection}, we use the Area Under the ROC Curve (i.e., AUC-ROC) as the metric, repeat the experiments four times, and report the performance of \name\ with baselines.

From Table~\ref{tb:anomaly detection}, we can observe that the detection module of \name\ achieves very competitive performance. An explanation is that, based on the anomalies distribution shown in Table~\ref{tb:label_distribution}, it can be observed that the anomalies are very rare events. Our generative manner could deal with the very rare scenario by learning the feature latent distributions instead of the (semi-)supervised learning manner.
For example, the maximum frequency of occurrences of thunderstorms is 770 (i.e., Jun 2017), which is collected from 238 counties over $30 \times 24 = 720$ hours, and the corresponding percentage is $\frac{770}{238 \times 30 \times 24} \approx 0.45\%$. 
Recall Remark~\ref{remark}, facing such rare events, we possibly find a single distribution to fit various anomaly patterns.

\section{Related Work}

\label{sec:related_work}
Noteworthy applications of graph learning techniques in time series forecasting span in climate domains, including but not limited to heatwave prediction~\cite{li2023regional}, and frost forecasts~\cite{lira2022graph}.
To improve the the time series analysis effectiveness, there has been a growing focus on structured learning in the context of tabular time series data~\cite{DBLP:conf/iclr/LiYS018, DBLP:conf/kdd/WuPL0CZ20, DBLP:conf/icdm/ZhaoWDHCTXBTZ20, DBLP:conf/nips/CaoWDZZHTXBTZ20, DBLP:conf/iclr/Shang0B21, DBLP:conf/aaai/DengH21, DBLP:conf/iclr/MarcinkevicsV21, DBLP:journals/corr/abs-2202-02195, DBLP:journals/pami/TankCFSF22, DBLP:journals/pami/SpadonHBMRS22, gong2023rhino}, which learned structures contribute to various time series analysis tasks like forecasting, anomaly detection, imputation, etc. As a directed and interpretable structure, causal graphs attract much research attention in this research topic~\cite{DBLP:journals/csur/GuoCLH020}.
Granger Causality is a classic tool for discovering the cross-time variable causality in time series~\cite{granger1969investigating, DBLP:conf/kdd/ArnoldLA07}.
Facing complex patterns in time series data, different upgraded Granger Causality discovery methods emerge in different directions. 
Also, neural Granger Causality tools are recently proposed~\cite{DBLP:journals/pami/TankCFSF22,DBLP:journals/make/NautaBS19,DBLP:conf/iclr/KhannaT20, DBLP:conf/iclr/MarcinkevicsV21, DBLP:conf/cikm/XuHY19, DBLP:conf/cikm/HuangXYYWX20}, which utilizes the deep neural network to discover the nonlinear Granger causal coefficients and serve for the time-series forecasting tasks better. For example, in~\cite{DBLP:journals/pami/TankCFSF22}, authors introduce how to use multi-layer perception (MLPs) and long short-term memory (LSTMs) to realize the Neural Granger Causality for the forecasting task and how to interpret the Granger causal coefficients from neurons in deep networks.
However, Granger Causality or Neural Granger Causality focuses on cross-time variable causality discovery and overlooks the instantaneous (or intra-time) variable causality.
Also, how to utilize the discovered comprehensive causality to contribute to the downstream time series analysis tasks is under-explored mainly, especially in a setting where the ground-truth causal structures are hardly available for evaluation.

\section{Conclusion}

In this paper, we first propose TBN Granger Causality to align the instantaneous causal effects with time-lagged Granger causality. Moreover, we design \name\ to use TBN Granger Causality on time series analysis tasks like forecasting and anomaly detection in the real-world tensor time-series data and perform extensive experiments, where the results show the effectiveness of \name.

\section*{Acknowledgement}
This work is supported by National Science Foundation under Award No. IIS-2117902, MIT-IBM Watson AI Lab, and IBM-Illinois Discovery Accelerator Institute - a new model of an academic-industry partnership designed to increase access to technology education and skill development to spur breakthroughs in emerging areas of technology. The views and conclusions are those of the authors and should not be interpreted as representing the official policies of the funding agencies or the government.




\nocite{langley00}

\bibliography{reference}

\begin{thebibliography}{54}
\providecommand{\natexlab}[1]{#1}
\providecommand{\url}[1]{\texttt{#1}}
\expandafter\ifx\csname urlstyle\endcsname\relax
  \providecommand{\doi}[1]{doi: #1}\else
  \providecommand{\doi}{doi: \begingroup \urlstyle{rm}\Url}\fi

\bibitem[Arnold et~al.(2007)Arnold, Liu, and Abe]{DBLP:conf/kdd/ArnoldLA07}
Arnold, A., Liu, Y., and Abe, N.
\newblock Temporal causal modeling with graphical granger methods.
\newblock In Berkhin, P., Caruana, R., and Wu, X. (eds.), \emph{Proceedings of the 13th {ACM} {SIGKDD} International Conference on Knowledge Discovery and Data Mining, San Jose, California, USA, August 12-15, 2007}, pp.\  66--75. {ACM}, 2007.
\newblock \doi{10.1145/1281192.1281203}.
\newblock URL \url{https://doi.org/10.1145/1281192.1281203}.

\bibitem[Assaad et~al.(2022)Assaad, Devijver, and Gaussier]{DBLP:conf/uai/AssaadDG22}
Assaad, C.~K., Devijver, E., and Gaussier, {\'{E}}.
\newblock Discovery of extended summary graphs in time series.
\newblock In Cussens, J. and Zhang, K. (eds.), \emph{Uncertainty in Artificial Intelligence, Proceedings of the Thirty-Eighth Conference on Uncertainty in Artificial Intelligence, {UAI} 2022, 1-5 August 2022, Eindhoven, The Netherlands}, volume 180 of \emph{Proceedings of Machine Learning Research}, pp.\  96--106. {PMLR}, 2022.
\newblock URL \url{https://proceedings.mlr.press/v180/assaad22a.html}.

\bibitem[Beirlant et~al.(2004)Beirlant, Goegebeur, Segers, and Teugels]{beirlant2004statistics}
Beirlant, J., Goegebeur, Y., Segers, J., and Teugels, J.~L.
\newblock \emph{Statistics of extremes: theory and applications}, volume 558.
\newblock John Wiley \& Sons, 2004.

\bibitem[Cao et~al.(2020)Cao, Wang, Duan, Zhang, Zhu, Huang, Tong, Xu, Bai, Tong, and Zhang]{DBLP:conf/nips/CaoWDZZHTXBTZ20}
Cao, D., Wang, Y., Duan, J., Zhang, C., Zhu, X., Huang, C., Tong, Y., Xu, B., Bai, J., Tong, J., and Zhang, Q.
\newblock Spectral temporal graph neural network for multivariate time-series forecasting.
\newblock In Larochelle, H., Ranzato, M., Hadsell, R., Balcan, M., and Lin, H. (eds.), \emph{Advances in Neural Information Processing Systems 33: Annual Conference on Neural Information Processing Systems 2020, NeurIPS 2020, December 6-12, 2020, virtual}, 2020.
\newblock URL \url{https://proceedings.neurips.cc/paper/2020/hash/cdf6581cb7aca4b7e19ef136c6e601a5-Abstract.html}.

\bibitem[Chung et~al.(2014)Chung, G{\"{u}}l{\c{c}}ehre, Cho, and Bengio]{DBLP:journals/corr/ChungGCB14}
Chung, J., G{\"{u}}l{\c{c}}ehre, {\c{C}}., Cho, K., and Bengio, Y.
\newblock Empirical evaluation of gated recurrent neural networks on sequence modeling.
\newblock \emph{CoRR}, abs/1412.3555, 2014.
\newblock URL \url{http://arxiv.org/abs/1412.3555}.

\bibitem[Dahlhaus \& Eichler(2003)Dahlhaus and Eichler]{dahlhaus2003causality}
Dahlhaus, R. and Eichler, M.
\newblock Causality and graphical models in time series analysis.
\newblock \emph{Oxford Statistical Science Series}, pp.\  115--137, 2003.

\bibitem[Deng \& Hooi(2021)Deng and Hooi]{DBLP:conf/aaai/DengH21}
Deng, A. and Hooi, B.
\newblock Graph neural network-based anomaly detection in multivariate time series.
\newblock In \emph{Thirty-Fifth {AAAI} Conference on Artificial Intelligence, {AAAI} 2021, Thirty-Third Conference on Innovative Applications of Artificial Intelligence, {IAAI} 2021, The Eleventh Symposium on Educational Advances in Artificial Intelligence, {EAAI} 2021, Virtual Event, February 2-9, 2021}, pp.\  4027--4035. {AAAI} Press, 2021.
\newblock \doi{10.1609/aaai.v35i5.16523}.
\newblock URL \url{https://doi.org/10.1609/aaai.v35i5.16523}.

\bibitem[Fisher \& Tippett(1928)Fisher and Tippett]{fisher1928limiting}
Fisher, R.~A. and Tippett, L. H.~C.
\newblock Limiting forms of the frequency distribution of the largest or smallest member of a sample.
\newblock In \emph{Mathematical proceedings of the Cambridge philosophical society}, volume~24, pp.\  180--190. Cambridge University Press, 1928.

\bibitem[Fu \& He(2022)Fu and He]{DBLP:journals/fdata/FuH22}
Fu, D. and He, J.
\newblock Natural and artificial dynamics in graphs: Concept, progress, and future.
\newblock \emph{Frontiers Big Data}, 5, 2022.
\newblock \doi{10.3389/FDATA.2022.1062637}.
\newblock URL \url{https://doi.org/10.3389/fdata.2022.1062637}.

\bibitem[Fu et~al.(2023)Fu, Xu, Tong, and He]{DBLP:conf/wsdm/FuXTH23}
Fu, D., Xu, Z., Tong, H., and He, J.
\newblock Natural and artificial dynamics in gnns: {A} tutorial.
\newblock In Chua, T., Lauw, H.~W., Si, L., Terzi, E., and Tsaparas, P. (eds.), \emph{Proceedings of the Sixteenth {ACM} International Conference on Web Search and Data Mining, {WSDM} 2023, Singapore, 27 February 2023 - 3 March 2023}, pp.\  1252--1255. {ACM}, 2023.
\newblock \doi{10.1145/3539597.3572726}.
\newblock URL \url{https://doi.org/10.1145/3539597.3572726}.

\bibitem[Fu et~al.(2024)Fu, Hua, Xie, Fang, Zhang, Sancak, Wu, Malevich, He, and Long]{DBLP:journals/corr/abs-2403-16030}
Fu, D., Hua, Z., Xie, Y., Fang, J., Zhang, S., Sancak, K., Wu, H., Malevich, A., He, J., and Long, B.
\newblock Vcr-graphormer: {A} mini-batch graph transformer via virtual connections.
\newblock \emph{CoRR}, abs/2403.16030, 2024.
\newblock \doi{10.48550/ARXIV.2403.16030}.
\newblock URL \url{https://doi.org/10.48550/arXiv.2403.16030}.

\bibitem[Geffner et~al.(2022)Geffner, Antor{\'{a}}n, Foster, Gong, Ma, Kiciman, Sharma, Lamb, Kukla, Pawlowski, Allamanis, and Zhang]{DBLP:journals/corr/abs-2202-02195}
Geffner, T., Antor{\'{a}}n, J., Foster, A., Gong, W., Ma, C., Kiciman, E., Sharma, A., Lamb, A., Kukla, M., Pawlowski, N., Allamanis, M., and Zhang, C.
\newblock Deep end-to-end causal inference.
\newblock \emph{CoRR}, abs/2202.02195, 2022.
\newblock URL \url{https://arxiv.org/abs/2202.02195}.

\bibitem[Gong et~al.(2023)Gong, Jennings, Zhang, and Pawlowski]{gong2023rhino}
Gong, W., Jennings, J., Zhang, C., and Pawlowski, N.
\newblock Rhino: Deep causal temporal relationship learning with history-dependent noise.
\newblock In \emph{The Eleventh International Conference on Learning Representations}, 2023.
\newblock URL \url{https://openreview.net/forum?id=i_1rbq8yFWC}.

\bibitem[Goyal et~al.(2020)Goyal, Raghunathan, Jain, Simhadri, and Jain]{DBLP:conf/icml/GoyalRJS020}
Goyal, S., Raghunathan, A., Jain, M., Simhadri, H.~V., and Jain, P.
\newblock {DROCC:} deep robust one-class classification.
\newblock In \emph{Proceedings of the 37th International Conference on Machine Learning, {ICML} 2020, 13-18 July 2020, Virtual Event}, volume 119 of \emph{Proceedings of Machine Learning Research}, pp.\  3711--3721. {PMLR}, 2020.
\newblock URL \url{http://proceedings.mlr.press/v119/goyal20c.html}.

\bibitem[Granger(1969)]{granger1969investigating}
Granger, C.~W.
\newblock Investigating causal relations by econometric models and cross-spectral methods.
\newblock \emph{Econometrica: journal of the Econometric Society}, pp.\  424--438, 1969.

\bibitem[Guo et~al.(2021)Guo, Cheng, Li, Hahn, and Liu]{DBLP:journals/csur/GuoCLH020}
Guo, R., Cheng, L., Li, J., Hahn, P.~R., and Liu, H.
\newblock A survey of learning causality with data: Problems and methods.
\newblock \emph{{ACM} Comput. Surv.}, 53\penalty0 (4):\penalty0 75:1--75:37, 2021.
\newblock \doi{10.1145/3397269}.
\newblock URL \url{https://doi.org/10.1145/3397269}.

\bibitem[Hersbach et~al.(2018)Hersbach, Bell, Berrisford, Biavati, Hor{\'a}nyi, Mu{\~n}oz~Sabater, Nicolas, Peubey, Radu, Rozum, et~al.]{hersbach2018era5}
Hersbach, H., Bell, B., Berrisford, P., Biavati, G., Hor{\'a}nyi, A., Mu{\~n}oz~Sabater, J., Nicolas, J., Peubey, C., Radu, R., Rozum, I., et~al.
\newblock Era5 hourly data on single levels from 1979 to present.
\newblock \emph{Copernicus climate change service (c3s) climate data store (cds)}, 10\penalty0 (10.24381), 2018.

\bibitem[Huang et~al.(2020)Huang, Xu, Yoo, Yan, Wang, and Xue]{DBLP:conf/cikm/HuangXYYWX20}
Huang, H., Xu, C., Yoo, S., Yan, W., Wang, T., and Xue, F.
\newblock Imbalanced time series classification for flight data analyzing with nonlinear granger causality learning.
\newblock In d'Aquin, M., Dietze, S., Hauff, C., Curry, E., and Cudr{\'{e}}{-}Mauroux, P. (eds.), \emph{{CIKM} '20: The 29th {ACM} International Conference on Information and Knowledge Management, Virtual Event, Ireland, October 19-23, 2020}, pp.\  2533--2540. {ACM}, 2020.
\newblock \doi{10.1145/3340531.3412710}.
\newblock URL \url{https://doi.org/10.1145/3340531.3412710}.

\bibitem[Jang et~al.(2017)Jang, Gu, and Poole]{DBLP:conf/iclr/JangGP17}
Jang, E., Gu, S., and Poole, B.
\newblock Categorical reparameterization with gumbel-softmax.
\newblock In \emph{5th International Conference on Learning Representations, {ICLR} 2017, Toulon, France, April 24-26, 2017, Conference Track Proceedings}. OpenReview.net, 2017.
\newblock URL \url{https://openreview.net/forum?id=rkE3y85ee}.

\bibitem[Ji et~al.(2023)Ji, Wang, Huang, Wu, Xu, Wu, Zhang, and Zheng]{DBLP:conf/aaai/JiW00XWZZ23}
Ji, J., Wang, J., Huang, C., Wu, J., Xu, B., Wu, Z., Zhang, J., and Zheng, Y.
\newblock Spatio-temporal self-supervised learning for traffic flow prediction.
\newblock In \emph{{AAAI} 2023}, 2023.

\bibitem[Khanna \& Tan(2020)Khanna and Tan]{DBLP:conf/iclr/KhannaT20}
Khanna, S. and Tan, V. Y.~F.
\newblock Economy statistical recurrent units for inferring nonlinear granger causality.
\newblock In \emph{8th International Conference on Learning Representations, {ICLR} 2020, Addis Ababa, Ethiopia, April 26-30, 2020}. OpenReview.net, 2020.
\newblock URL \url{https://openreview.net/forum?id=SyxV9ANFDH}.

\bibitem[Kingma \& Welling(2014)Kingma and Welling]{DBLP:journals/corr/KingmaW13}
Kingma, D.~P. and Welling, M.
\newblock Auto-encoding variational bayes.
\newblock In Bengio, Y. and LeCun, Y. (eds.), \emph{2nd International Conference on Learning Representations, {ICLR} 2014, Banff, AB, Canada, April 14-16, 2014, Conference Track Proceedings}, 2014.
\newblock URL \url{http://arxiv.org/abs/1312.6114}.

\bibitem[Kipf \& Welling(2016)Kipf and Welling]{DBLP:journals/corr/KipfW16a}
Kipf, T.~N. and Welling, M.
\newblock Variational graph auto-encoders.
\newblock \emph{CoRR}, abs/1611.07308, 2016.
\newblock URL \url{http://arxiv.org/abs/1611.07308}.

\bibitem[Kipf et~al.(2018)Kipf, Fetaya, Wang, Welling, and Zemel]{DBLP:conf/icml/KipfFWWZ18}
Kipf, T.~N., Fetaya, E., Wang, K., Welling, M., and Zemel, R.~S.
\newblock Neural relational inference for interacting systems.
\newblock In Dy, J.~G. and Krause, A. (eds.), \emph{Proceedings of the 35th International Conference on Machine Learning, {ICML} 2018, Stockholmsm{\"{a}}ssan, Stockholm, Sweden, July 10-15, 2018}, volume~80 of \emph{Proceedings of Machine Learning Research}, pp.\  2693--2702. {PMLR}, 2018.
\newblock URL \url{http://proceedings.mlr.press/v80/kipf18a.html}.

\bibitem[Kofinas et~al.(2021)Kofinas, Nagaraja, and Gavves]{DBLP:conf/nips/KofinasNG21}
Kofinas, M., Nagaraja, N.~S., and Gavves, E.
\newblock Roto-translated local coordinate frames for interacting dynamical systems.
\newblock In \emph{NeurIPS 2021}, 2021.

\bibitem[Kofinas et~al.(2023)Kofinas, Bekkers, Nagaraja, and Gavves]{DBLP:journals/corr/abs-2310-20679}
Kofinas, M., Bekkers, E.~J., Nagaraja, N.~S., and Gavves, E.
\newblock Latent field discovery in interacting dynamical systems with neural fields.
\newblock \emph{CoRR}, abs/2310.20679, 2023.
\newblock \doi{10.48550/ARXIV.2310.20679}.
\newblock URL \url{https://doi.org/10.48550/arXiv.2310.20679}.

\bibitem[Li et~al.(2023{\natexlab{a}})Li, Yu, Huang, Wang, and Sharma]{li2023regional}
Li, P., Yu, Y., Huang, D., Wang, Z.-H., and Sharma, A.
\newblock Regional heatwave prediction using graph neural network and weather station data.
\newblock \emph{Geophysical Research Letters}, 50\penalty0 (7):\penalty0 e2023GL103405, 2023{\natexlab{a}}.

\bibitem[Li et~al.(2018)Li, Yu, Shahabi, and Liu]{DBLP:conf/iclr/LiYS018}
Li, Y., Yu, R., Shahabi, C., and Liu, Y.
\newblock Diffusion convolutional recurrent neural network: Data-driven traffic forecasting.
\newblock In \emph{6th International Conference on Learning Representations, {ICLR} 2018, Vancouver, BC, Canada, April 30 - May 3, 2018, Conference Track Proceedings}. OpenReview.net, 2018.
\newblock URL \url{https://openreview.net/forum?id=SJiHXGWAZ}.

\bibitem[Li et~al.(2023{\natexlab{b}})Li, Fu, and He]{DBLP:conf/www/LiFH23}
Li, Z., Fu, D., and He, J.
\newblock Everything evolves in personalized pagerank.
\newblock In Ding, Y., Tang, J., Sequeda, J.~F., Aroyo, L., Castillo, C., and Houben, G. (eds.), \emph{Proceedings of the {ACM} Web Conference 2023, {WWW} 2023, Austin, TX, USA, 30 April 2023 - 4 May 2023}, pp.\  3342--3352. {ACM}, 2023{\natexlab{b}}.
\newblock \doi{10.1145/3543507.3583474}.
\newblock URL \url{https://doi.org/10.1145/3543507.3583474}.

\bibitem[Lira et~al.(2022)Lira, Mart{\'\i}, and Sanchez-Pi]{lira2022graph}
Lira, H., Mart{\'\i}, L., and Sanchez-Pi, N.
\newblock A graph neural network with spatio-temporal attention for multi-sources time series data: An application to frost forecast.
\newblock \emph{Sensors}, 22\penalty0 (4):\penalty0 1486, 2022.

\bibitem[Lorenz(1996)]{lorenz1996predictability}
Lorenz, E.~N.
\newblock Predictability: A problem partly solved.
\newblock In \emph{Proc. Seminar on predictability}, volume~1. Reading, 1996.

\bibitem[Maddison et~al.(2017)Maddison, Mnih, and Teh]{DBLP:conf/iclr/MaddisonMT17}
Maddison, C.~J., Mnih, A., and Teh, Y.~W.
\newblock The concrete distribution: {A} continuous relaxation of discrete random variables.
\newblock In \emph{5th International Conference on Learning Representations, {ICLR} 2017, Toulon, France, April 24-26, 2017, Conference Track Proceedings}. OpenReview.net, 2017.
\newblock URL \url{https://openreview.net/forum?id=S1jE5L5gl}.

\bibitem[Malinsky \& Spirtes(2018)Malinsky and Spirtes]{DBLP:conf/kdd/MalinskyS18}
Malinsky, D. and Spirtes, P.
\newblock Causal structure learning from multivariate time series in settings with unmeasured confounding.
\newblock In Le, T.~D., Zhang, K., Kiciman, E., Hyv{\"{a}}rinen, A., and Liu, L. (eds.), \emph{Proceedings of 2018 {ACM} {SIGKDD} Workshop on Causal Discovery, CD@KDD 2018, London, UK, 20 August 2018}, volume~92 of \emph{Proceedings of Machine Learning Research}, pp.\  23--47. {PMLR}, 2018.
\newblock URL \url{http://proceedings.mlr.press/v92/malinsky18a.html}.

\bibitem[Marcinkevics \& Vogt(2021)Marcinkevics and Vogt]{DBLP:conf/iclr/MarcinkevicsV21}
Marcinkevics, R. and Vogt, J.~E.
\newblock Interpretable models for granger causality using self-explaining neural networks.
\newblock In \emph{9th International Conference on Learning Representations, {ICLR} 2021, Virtual Event, Austria, May 3-7, 2021}. OpenReview.net, 2021.
\newblock URL \url{https://openreview.net/forum?id=DEa4JdMWRHp}.

\bibitem[Moneta et~al.(2013)Moneta, Entner, Hoyer, and Coad]{moneta2013causal}
Moneta, A., Entner, D., Hoyer, P.~O., and Coad, A.
\newblock Causal inference by independent component analysis: Theory and applications.
\newblock \emph{Oxford Bulletin of Economics and Statistics}, 75\penalty0 (5):\penalty0 705--730, 2013.

\bibitem[Nauta et~al.(2019)Nauta, Bucur, and Seifert]{DBLP:journals/make/NautaBS19}
Nauta, M., Bucur, D., and Seifert, C.
\newblock Causal discovery with attention-based convolutional neural networks.
\newblock \emph{Mach. Learn. Knowl. Extr.}, 1\penalty0 (1):\penalty0 312--340, 2019.
\newblock \doi{10.3390/make1010019}.
\newblock URL \url{https://doi.org/10.3390/make1010019}.

\bibitem[Pamfil et~al.(2020)Pamfil, Sriwattanaworachai, Desai, Pilgerstorfer, Georgatzis, Beaumont, and Aragam]{DBLP:conf/aistats/PamfilSDPGBA20}
Pamfil, R., Sriwattanaworachai, N., Desai, S., Pilgerstorfer, P., Georgatzis, K., Beaumont, P., and Aragam, B.
\newblock {DYNOTEARS:} structure learning from time-series data.
\newblock In Chiappa, S. and Calandra, R. (eds.), \emph{The 23rd International Conference on Artificial Intelligence and Statistics, {AISTATS} 2020, 26-28 August 2020, Online [Palermo, Sicily, Italy]}, volume 108 of \emph{Proceedings of Machine Learning Research}, pp.\  1595--1605. {PMLR}, 2020.
\newblock URL \url{http://proceedings.mlr.press/v108/pamfil20a.html}.

\bibitem[Ruff et~al.(2018)Ruff, G{\"{o}}rnitz, Deecke, Siddiqui, Vandermeulen, Binder, M{\"{u}}ller, and Kloft]{DBLP:conf/icml/RuffGDSVBMK18}
Ruff, L., G{\"{o}}rnitz, N., Deecke, L., Siddiqui, S.~A., Vandermeulen, R.~A., Binder, A., M{\"{u}}ller, E., and Kloft, M.
\newblock Deep one-class classification.
\newblock In Dy, J.~G. and Krause, A. (eds.), \emph{Proceedings of the 35th International Conference on Machine Learning, {ICML} 2018, Stockholmsm{\"{a}}ssan, Stockholm, Sweden, July 10-15, 2018}, volume~80 of \emph{Proceedings of Machine Learning Research}, pp.\  4390--4399. {PMLR}, 2018.
\newblock URL \url{http://proceedings.mlr.press/v80/ruff18a.html}.

\bibitem[Ruff et~al.(2020)Ruff, Vandermeulen, G{\"{o}}rnitz, Binder, M{\"{u}}ller, M{\"{u}}ller, and Kloft]{DBLP:conf/iclr/RuffVGBMMK20}
Ruff, L., Vandermeulen, R.~A., G{\"{o}}rnitz, N., Binder, A., M{\"{u}}ller, E., M{\"{u}}ller, K., and Kloft, M.
\newblock Deep semi-supervised anomaly detection.
\newblock In \emph{8th International Conference on Learning Representations, {ICLR} 2020, Addis Ababa, Ethiopia, April 26-30, 2020}. OpenReview.net, 2020.
\newblock URL \url{https://openreview.net/forum?id=HkgH0TEYwH}.

\bibitem[Runge et~al.(2019)Runge, Nowack, Kretschmer, Flaxman, and Sejdinovic]{runge2019detecting}
Runge, J., Nowack, P., Kretschmer, M., Flaxman, S., and Sejdinovic, D.
\newblock Detecting and quantifying causal associations in large nonlinear time series datasets.
\newblock \emph{Science advances}, 5\penalty0 (11):\penalty0 eaau4996, 2019.

\bibitem[Shang et~al.(2021)Shang, Chen, and Bi]{DBLP:conf/iclr/Shang0B21}
Shang, C., Chen, J., and Bi, J.
\newblock Discrete graph structure learning for forecasting multiple time series.
\newblock In \emph{9th International Conference on Learning Representations, {ICLR} 2021, Virtual Event, Austria, May 3-7, 2021}. OpenReview.net, 2021.
\newblock URL \url{https://openreview.net/forum?id=WEHSlH5mOk}.

\bibitem[Siffer et~al.(2017)Siffer, Fouque, Termier, and Largou{\"{e}}t]{DBLP:conf/kdd/SifferFTL17}
Siffer, A., Fouque, P., Termier, A., and Largou{\"{e}}t, C.
\newblock Anomaly detection in streams with extreme value theory.
\newblock In \emph{Proceedings of the 23rd {ACM} {SIGKDD} International Conference on Knowledge Discovery and Data Mining, Halifax, NS, Canada, August 13 - 17, 2017}, pp.\  1067--1075. {ACM}, 2017.
\newblock \doi{10.1145/3097983.3098144}.
\newblock URL \url{https://doi.org/10.1145/3097983.3098144}.

\bibitem[Spadon et~al.(2022)Spadon, Hong, Brandoli, Matwin, Jr., and Sun]{DBLP:journals/pami/SpadonHBMRS22}
Spadon, G., Hong, S., Brandoli, B., Matwin, S., Jr., J. F.~R., and Sun, J.
\newblock Pay attention to evolution: Time series forecasting with deep graph-evolution learning.
\newblock \emph{{IEEE} Trans. Pattern Anal. Mach. Intell.}, 44\penalty0 (9):\penalty0 5368--5384, 2022.
\newblock \doi{10.1109/TPAMI.2021.3076155}.
\newblock URL \url{https://doi.org/10.1109/TPAMI.2021.3076155}.

\bibitem[Su et~al.(2019)Su, Zhao, Niu, Liu, Sun, and Pei]{DBLP:conf/kdd/SuZNLSP19}
Su, Y., Zhao, Y., Niu, C., Liu, R., Sun, W., and Pei, D.
\newblock Robust anomaly detection for multivariate time series through stochastic recurrent neural network.
\newblock In Teredesai, A., Kumar, V., Li, Y., Rosales, R., Terzi, E., and Karypis, G. (eds.), \emph{Proceedings of the 25th {ACM} {SIGKDD} International Conference on Knowledge Discovery {\&} Data Mining, {KDD} 2019, Anchorage, AK, USA, August 4-8, 2019}, pp.\  2828--2837. {ACM}, 2019.
\newblock \doi{10.1145/3292500.3330672}.
\newblock URL \url{https://doi.org/10.1145/3292500.3330672}.

\bibitem[Sutskever et~al.(2014)Sutskever, Vinyals, and Le]{DBLP:conf/nips/SutskeverVL14}
Sutskever, I., Vinyals, O., and Le, Q.~V.
\newblock Sequence to sequence learning with neural networks.
\newblock In Ghahramani, Z., Welling, M., Cortes, C., Lawrence, N.~D., and Weinberger, K.~Q. (eds.), \emph{Advances in Neural Information Processing Systems 27: Annual Conference on Neural Information Processing Systems 2014, December 8-13 2014, Montreal, Quebec, Canada}, pp.\  3104--3112, 2014.
\newblock URL \url{https://proceedings.neurips.cc/paper/2014/hash/a14ac55a4f27472c5d894ec1c3c743d2-Abstract.html}.

\bibitem[Tank et~al.(2022)Tank, Covert, Foti, Shojaie, and Fox]{DBLP:journals/pami/TankCFSF22}
Tank, A., Covert, I., Foti, N.~J., Shojaie, A., and Fox, E.~B.
\newblock Neural granger causality.
\newblock \emph{{IEEE} Trans. Pattern Anal. Mach. Intell.}, 44\penalty0 (8):\penalty0 4267--4279, 2022.
\newblock \doi{10.1109/TPAMI.2021.3065601}.
\newblock URL \url{https://doi.org/10.1109/TPAMI.2021.3065601}.

\bibitem[Wild et~al.(2010)Wild, Eichler, Friederich, Hartmann, Zipfel, and Herzog]{wild2010graphical}
Wild, B., Eichler, M., Friederich, H.-C., Hartmann, M., Zipfel, S., and Herzog, W.
\newblock A graphical vector autoregressive modelling approach to the analysis of electronic diary data.
\newblock \emph{BMC medical research methodology}, 10:\penalty0 1--13, 2010.

\bibitem[Wu et~al.(2020)Wu, Pan, Long, Jiang, Chang, and Zhang]{DBLP:conf/kdd/WuPL0CZ20}
Wu, Z., Pan, S., Long, G., Jiang, J., Chang, X., and Zhang, C.
\newblock Connecting the dots: Multivariate time series forecasting with graph neural networks.
\newblock In Gupta, R., Liu, Y., Tang, J., and Prakash, B.~A. (eds.), \emph{{KDD} '20: The 26th {ACM} {SIGKDD} Conference on Knowledge Discovery and Data Mining, Virtual Event, CA, USA, August 23-27, 2020}, pp.\  753--763. {ACM}, 2020.
\newblock \doi{10.1145/3394486.3403118}.
\newblock URL \url{https://doi.org/10.1145/3394486.3403118}.

\bibitem[Wu et~al.(2021)Wu, Pan, Chen, Long, Zhang, and Yu]{DBLP:journals/tnn/WuPCLZY21}
Wu, Z., Pan, S., Chen, F., Long, G., Zhang, C., and Yu, P.~S.
\newblock A comprehensive survey on graph neural networks.
\newblock \emph{{IEEE} Trans. Neural Networks Learn. Syst.}, 32\penalty0 (1):\penalty0 4--24, 2021.
\newblock \doi{10.1109/TNNLS.2020.2978386}.
\newblock URL \url{https://doi.org/10.1109/TNNLS.2020.2978386}.

\bibitem[Xu et~al.(2019)Xu, Huang, and Yoo]{DBLP:conf/cikm/XuHY19}
Xu, C., Huang, H., and Yoo, S.
\newblock Scalable causal graph learning through a deep neural network.
\newblock In Zhu, W., Tao, D., Cheng, X., Cui, P., Rundensteiner, E.~A., Carmel, D., He, Q., and Yu, J.~X. (eds.), \emph{Proceedings of the 28th {ACM} International Conference on Information and Knowledge Management, {CIKM} 2019, Beijing, China, November 3-7, 2019}, pp.\  1853--1862. {ACM}, 2019.
\newblock \doi{10.1145/3357384.3357864}.
\newblock URL \url{https://doi.org/10.1145/3357384.3357864}.

\bibitem[Yu et~al.(2019)Yu, Chen, Gao, and Yu]{DBLP:conf/icml/YuCGY19}
Yu, Y., Chen, J., Gao, T., and Yu, M.
\newblock {DAG-GNN:} {DAG} structure learning with graph neural networks.
\newblock In Chaudhuri, K. and Salakhutdinov, R. (eds.), \emph{Proceedings of the 36th International Conference on Machine Learning, {ICML} 2019, 9-15 June 2019, Long Beach, California, {USA}}, volume~97 of \emph{Proceedings of Machine Learning Research}, pp.\  7154--7163. {PMLR}, 2019.
\newblock URL \url{http://proceedings.mlr.press/v97/yu19a.html}.

\bibitem[Zhao et~al.(2020)Zhao, Wang, Duan, Huang, Cao, Tong, Xu, Bai, Tong, and Zhang]{DBLP:conf/icdm/ZhaoWDHCTXBTZ20}
Zhao, H., Wang, Y., Duan, J., Huang, C., Cao, D., Tong, Y., Xu, B., Bai, J., Tong, J., and Zhang, Q.
\newblock Multivariate time-series anomaly detection via graph attention network.
\newblock In Plant, C., Wang, H., Cuzzocrea, A., Zaniolo, C., and Wu, X. (eds.), \emph{20th {IEEE} International Conference on Data Mining, {ICDM} 2020, Sorrento, Italy, November 17-20, 2020}, pp.\  841--850. {IEEE}, 2020.
\newblock \doi{10.1109/ICDM50108.2020.00093}.
\newblock URL \url{https://doi.org/10.1109/ICDM50108.2020.00093}.

\bibitem[Zheng et~al.(2018)Zheng, Aragam, Ravikumar, and Xing]{DBLP:conf/nips/ZhengARX18}
Zheng, X., Aragam, B., Ravikumar, P., and Xing, E.~P.
\newblock Dags with {NO} {TEARS:} continuous optimization for structure learning.
\newblock In Bengio, S., Wallach, H.~M., Larochelle, H., Grauman, K., Cesa{-}Bianchi, N., and Garnett, R. (eds.), \emph{Advances in Neural Information Processing Systems 31: Annual Conference on Neural Information Processing Systems 2018, NeurIPS 2018, December 3-8, 2018, Montr{\'{e}}al, Canada}, pp.\  9492--9503, 2018.
\newblock URL \url{https://proceedings.neurips.cc/paper/2018/hash/e347c51419ffb23ca3fd5050202f9c3d-Abstract.html}.

\bibitem[Zhou et~al.(2022)Zhou, Zheng, Fu, Han, and He]{DBLP:conf/cikm/ZhouZF0H22}
Zhou, D., Zheng, L., Fu, D., Han, J., and He, J.
\newblock Mentorgnn: Deriving curriculum for pre-training gnns.
\newblock In Hasan, M.~A. and Xiong, L. (eds.), \emph{Proceedings of the 31st {ACM} International Conference on Information {\&} Knowledge Management, Atlanta, GA, USA, October 17-21, 2022}, pp.\  2721--2731. {ACM}, 2022.
\newblock \doi{10.1145/3511808.3557393}.
\newblock URL \url{https://doi.org/10.1145/3511808.3557393}.

\end{thebibliography}
\bibliographystyle{icml2024}

\newpage
\appendix
\onecolumn

\section{Theoretical Analysis}
\label{sec:theoretical_analysis}

\subsection{Proof of Lemma~\ref{eq:acyclicity}}
Following~\cite{DBLP:conf/icml/YuCGY19}, at each time $t$, we can extend
$(\bm{I} + \bm{A}^{(t)} \circ \bm{A}^{(t)})^{N}$ by binomial expansion as follows.

\begin{equation}
    (\bm{I} + \bm{A}^{(t)} \circ \bm{A}^{(t)})^{N} = \bm{I} + \sum_{k=1}^{N} \begin{pmatrix} N \\ k \end{pmatrix} (\bm{A}^{(t)})^{k}
\end{equation}

Since 
\begin{equation}
    \bm{I} \in \mathbb{R}^{N \times N}
\end{equation}
then
\begin{equation}
    \text{Tr}(\bm{I}) = N
\end{equation}

Thus, if
\begin{equation}
    (\bm{I} + \bm{A}^{(t)} \circ \bm{A}^{(t)})^{N} - N = 0
\end{equation}
then
\begin{equation}
    (\bm{A}^{(t)})^{k} = 0, \text{~for any~} k
\end{equation}

Therefore, $\bm{A}^{(t)}$ is acyclic, i.e., no closed-loop exists in $\bm{A}^{(t)}$ at any possible length. Overall, the general idea of Lemma~\ref{eq:acyclicity} is to ensure that the diagonal entries of the powered adjacency matrix have no $1$s. There are also other forms for acyclicity constraints obeying the same idea but in different expressions, like exponential power form in~\cite{DBLP:conf/nips/ZhengARX18}.

\subsection{Sketch Proof of Theorem~\ref{eq: discovery}}
According to Theorem 1 from~\cite{DBLP:journals/corr/abs-2202-02195}, the ELBO form as our Eq.~\ref{eq: inner_opt} could identity the ground-truth causal structure $\mathbf{S}^{(t)}$ at each time $t$. The difference between our ELBO and the ELBO in~\cite{DBLP:journals/corr/abs-2202-02195} is entries in the KL-divergence. Specifically, in~\cite{DBLP:journals/corr/abs-2202-02195}, the prior and variational posterior distributions are on the graph level. Usually, the prior distribution of graph structures is not easy to obtain (e.g., the non-IID and heterophyllous properties). Then, we transfer the graph structure distribution to the feature distribution that the Gaussian distribution can model. That's why our prior and variational posterior distributions in the KL-divergence are on the feature (generated by the graph) level.

\section{Empirical Analysis}
\label{sec:empirical_analysis}

\subsection{Ground-Truth Causality Discovery Ability of \name}
Lorenz-96 model~\cite{lorenz1996predictability} is a famous synthetic system of multivariate time-series, e.g., $\bm{X} \in \mathbb{R}^{P \times T}$ is a $P$-dimensional time series whose dynamics can be modeled as follows.
\begin{equation}
    \frac{d\bm{X}(i,t)}{dt} = (\bm{X}(i+1,t) - \bm{X}(i-2,t))\bm{X}(i-1,t) - \bm{X}(i,t) + F, \text{~for~} i \in \{1, 2, \ldots, P\}
\end{equation}
where $\bm{X}(0,t) = \bm{X}(P,t)$, $\bm{X}(-1,t) = \bm{X}(P-1,t)$, $\bm{X}(P+1,t) = \bm{X}(1,t)$, and $F$ is the forcing constant determining the level of nonlinearity and chaos in the time series. With the above modeling, the corresponding ground-truth Granger causal structures can be simulated, involving multivariate, nonlinear, and sparse~\cite{DBLP:journals/pami/TankCFSF22}.

\begin{figure}[h]
    \centering    
    \begin{subfigure}[b]{0.45\textwidth}
        \includegraphics[width=\textwidth]{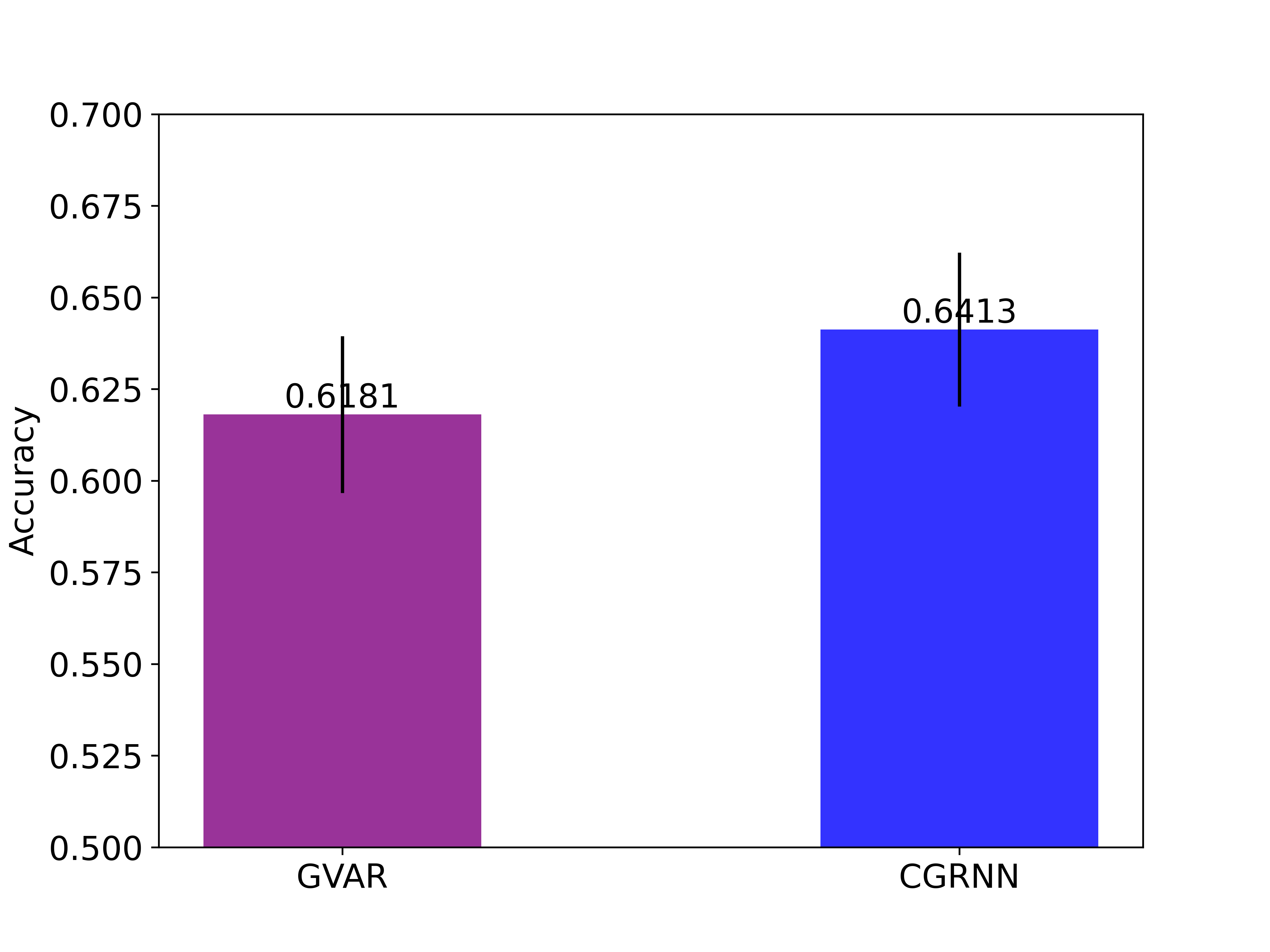}
        \caption{$P$=10, $T$=500}
    \end{subfigure}
    \begin{subfigure}[b]{0.45\textwidth}
        \includegraphics[width=\textwidth]{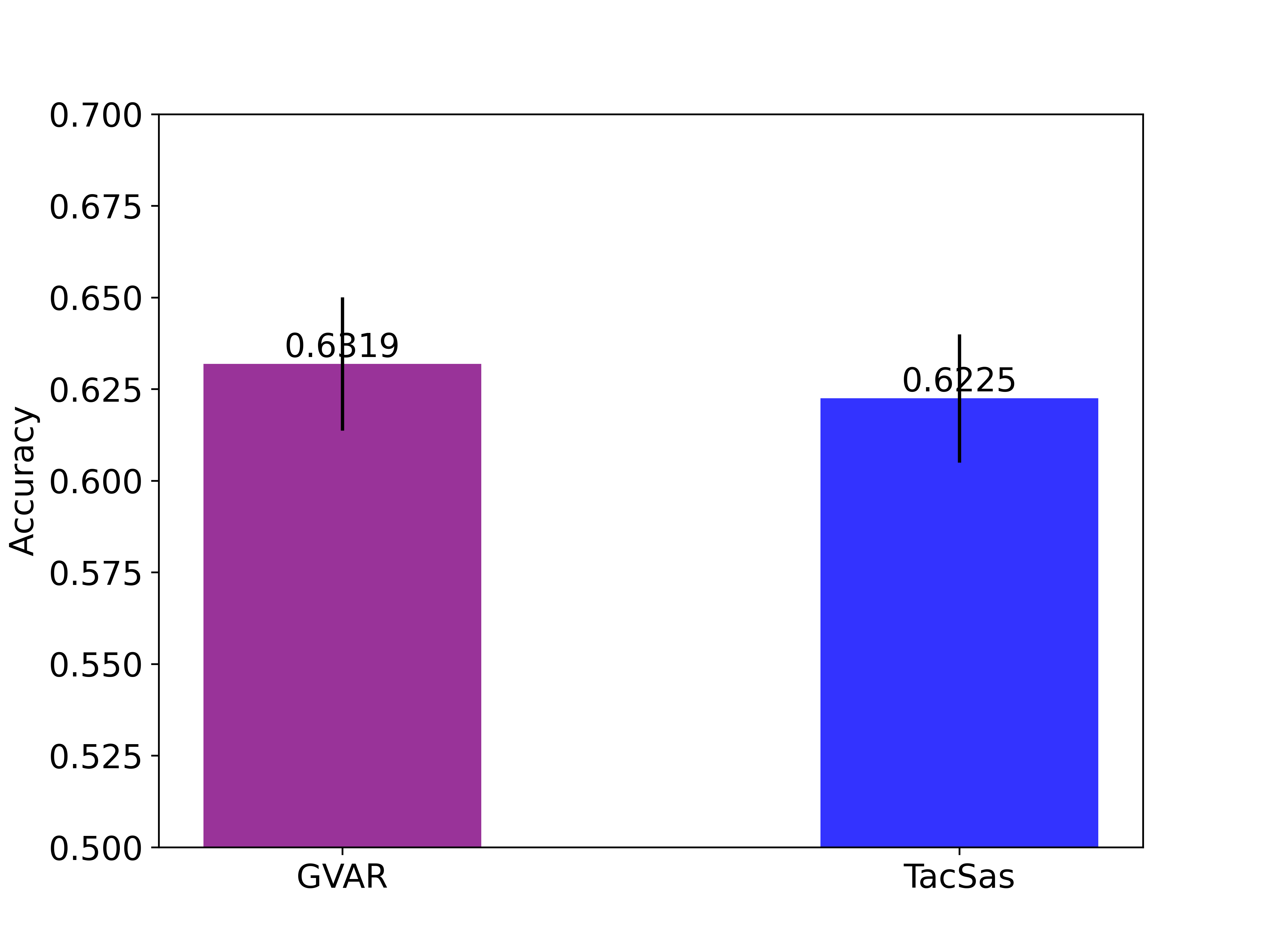}
        \caption{$P$=10, $T$=800}
    \end{subfigure}
    \begin{subfigure}[b]{0.45\textwidth}
        \includegraphics[width=\textwidth]{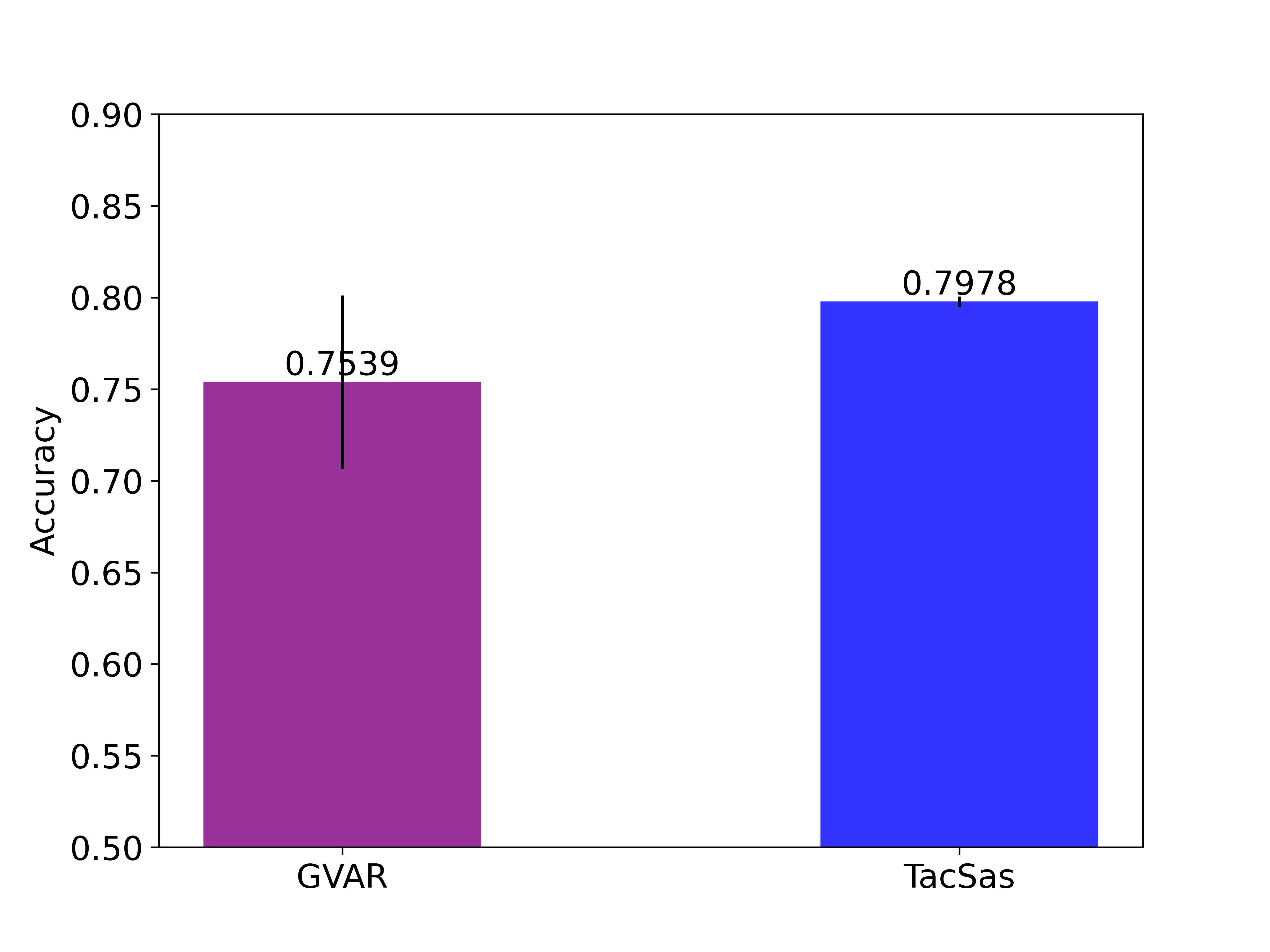}
        \caption{$P$=20, $T$=500}
    \end{subfigure}
    \begin{subfigure}[b]{0.45\textwidth}
        \includegraphics[width=\textwidth]{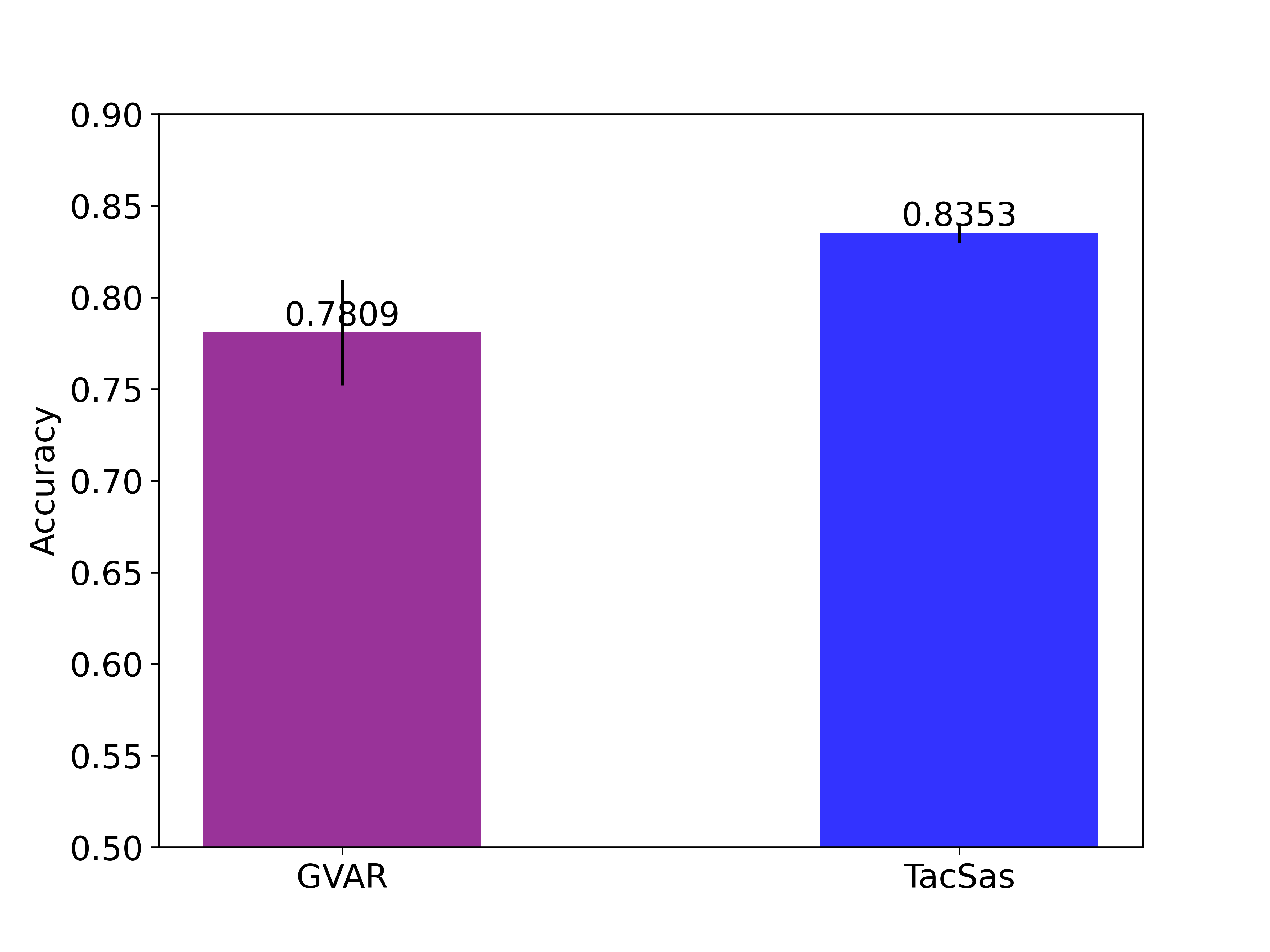}
        \caption{$P$=20, $T$=800}
    \end{subfigure}
\caption{Accuracy of Causality Discovery in Lorenz-96 with Varying Number of Variables ($P$) and Timestamps ($T$).}
\label{fig: causality_discovery}
\vspace{-7mm}
\end{figure}

To generate the ground-truth causal structures, there are two parameters, i.e., the number of variables (i.e., $P$) and the number of timestamps (i.e., $T$). Therefore, we control these two parameters and report the accuracy of \name\ discovered causal structures against the ground-truth ones (i.e., 0/1 adjacency matrices), compared with the state-of-the-art causality discovery method GVAR~\cite{DBLP:conf/iclr/MarcinkevicsV21}. The comparison is shown in Figure~\ref{fig: causality_discovery} after eight experiment trials with mean and variance computed, where we can observe our \name\ achieve the competitive accuracy of discovering the ground-truth causal structures. Also, by comparing Figure~\ref{fig: causality_discovery}(a) and (b) (and Figure~\ref{fig: causality_discovery}(c) and (d)), we can see that fixing the number of variables (i.e., $P$), increasing the time series length (i.e., $T$) may help discover the causality. And by comparing Figure~\ref{fig: causality_discovery}(a) and (c) (and Figure~\ref{fig: causality_discovery}(b) and (d)), we can see that fixing the time length (i.e., $T$), increasing the number of variables (i.e., $P$) may make the causality easier to be discovered.

\subsection{Validation of Anomaly Detection Ability of \name}

\begin{wrapfigure}{r}{0.45\textwidth}
\vspace{-7mm}
\begin{center}
    \includegraphics[width=0.41\textwidth]{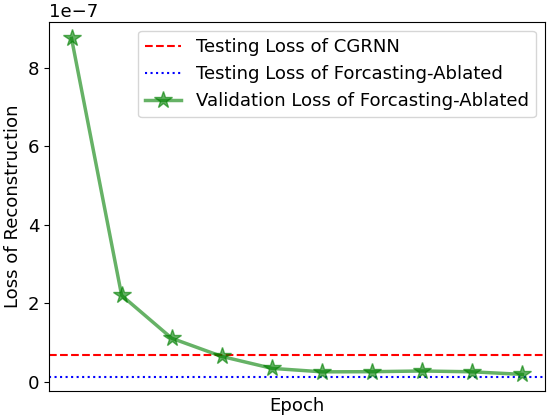}
  \end{center}
\vspace{-3mm}
\caption{Ablation of \name\ on Cross-Validation Group \#2 (i.e., 2018 as testing) }
\label{Fig:ablation}
\end{wrapfigure}
Besides forecasting, another capability of \name\ is anomaly detection. Based on the analysis of Remark~\ref{remark}, the detection function of \name\ originates from the accurate expression of the feature distribution. Although our forecast features have better accuracy than selected baselines (e.g., DCGNN and GTS), we need to verify if the forecast features still have a negligible divergence from the ground-truth features in terms of distribution. If so, we can safely use the forecast features to detect anomalies.
Therefore, we design the ablation study. We remove the forecasting part of \name\, i.e., we let the encoder and decoder in Figure~\ref{Fig:framework} directly learn the distribution of ground-truth features (instead of forecast features) and then test reconstruction loss on ground-truth features. In Figure~\ref{Fig:ablation}, we show the feature reconstruction loss (i.e., mean squared error) curve of the encoder and decoder on the validation set as the epoch increases. After the training of the encoder and decoder is converged, we can also observe that the ground-truth feature reconstruction loss does not have a very large divergence from the forecast features. Now, we are ready to do the following anomaly detection experiments.

\subsection{Ablation Study}
\begin{figure}[h]
\includegraphics[width=0.5\textwidth]{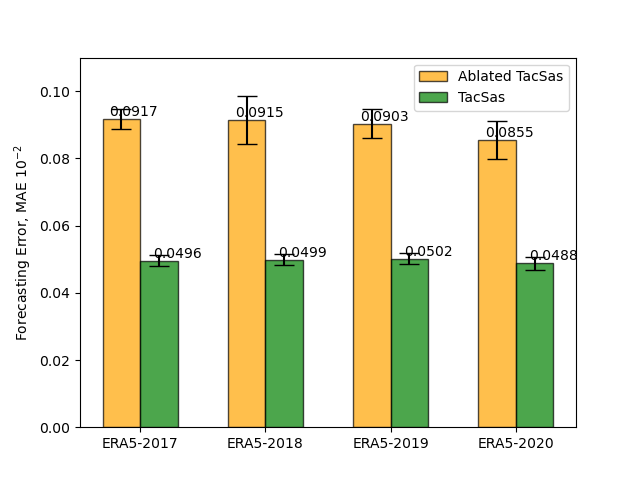}
\centering
\caption{Ablation Study on Forecasting Task.}
\label{Fig:abs}
\end{figure}

As shown in Table~\ref{tb:forecasting}, the GRU~\cite{DBLP:journals/corr/ChungGCB14} method does not perform well. A latent reason is that it can not take any structural information from the time series. Motivated by this guess, we designed the following ablation study on the forecasting task. The ablated \name\ is designed by only keeping the forget gate in Eq.~\ref{eq: gates}, i.e., the last equation in Eq.~\ref{eq: gates}, then all the rest of the gates follow the GRU method. As shown in Figure~\ref{Fig:abs}, we can see that only taking partial time-respecting causal structural information could not enable \name\ to achieve the best performance, but accepting this partial information can help GRU improve the performance compared with Table~\ref{tb:forecasting}.

\section{Implementation}
\label{sec:implementation}

\subsection{Hyperparameter Search}
In Eq.~\ref{eq: complete_inner_opt}, instead of fixing the hyperparameter $\lambda$ and $c$ during the optimization. Increasing the values of hyperparameter $\lambda$ and $c$ can reduce the possibility that learned structures break the acyclicity~\cite{DBLP:conf/icml/YuCGY19}, such that one iterative way to increase hyperparameters $\lambda$ and $c$ during the optimization can be expressed as follows.
\begin{equation}
    \lambda_{i+1} \leftarrow \lambda_{i} + c_{i} \alpha(\bm{A}^{(t)}_{i})
\end{equation}
and
\begin{equation}  
    c_{i+1} = \begin{cases}
    \eta c_{i} &\text {if $|\alpha(\bm{A}^{(t)}_{i})| > \gamma |\alpha(\bm{A}^{(t)}_{i-1})|$}\\
    c_{i} &\text {otherwise}
    \end{cases}
\end{equation}
where $\eta > 1$ and $0 < \gamma < 1$ are two hyperparameters, the condition $|\alpha(\bm{A}^{(t)}_{i})| > \gamma |\alpha(\bm{A}^{(t)}_{i-1})|$ means that the current acyclicity $\alpha(\bm{A}^{(t)}_{i})$ at the $i$-th iteration is not ideal, because it is not decreased below the $\gamma$ portion of $\alpha(\bm{A}^{(t)}_{i-1})$ from the last iteration $i-1$.

\subsection{Reproducibility}
For forecasting and anomaly detection, we have four cross-validation groups. For example, focusing on an interesting time interval each year (e.g., from May to August is the season for frequent thunderstorms), we set group \#1 with [2018, 2019, 2020] as training, [2021] as validation, and [2017] as testing. Thus, we have 8856 hours, 45 weather features, and 238 counties in the training set. The rest three groups are \{[2019, 2020, 2021], [2017], [2018]\}, \{[2020, 2021, 2017], [2018], [2019]\}, and \{[2021, 2017, 2018], [2019], [2020]\}, respectively. Therefore, \name\ and baselines are required to forecast the testing set and detect the anomaly patterns in the testing set.

The persistence forecasting can be expressed as 
\begin{equation}
    \bm{X}^{(t)}_{\name++} = \alpha \bm{X}^{(t)}_{\name} + (1-\alpha) \bm{X}^{(t-\tau)} ~\text {~s.t.~} \bm{X}^{(t)}_{\name} = \text{TacSas}(\bm{X}^{(t-\tau)})
\end{equation}
where $\tau$ is the time window, for example, in the experiments, $\tau$ = 24h.

\name\ is published~\footnote{\url{https://github.com/DongqiFu/TacSas}}. The experiments are programmed based on Python and Pytorch on a Windows machine with 64GB RAM and a 16GB RTX 5000 GPU.

\section{Tensor Time Series Dataset}
\label{sec:dataset}

\subsection{Geographic Distribution of the Time Series Data}
\begin{figure*}[h]
\includegraphics[width=0.8\textwidth]{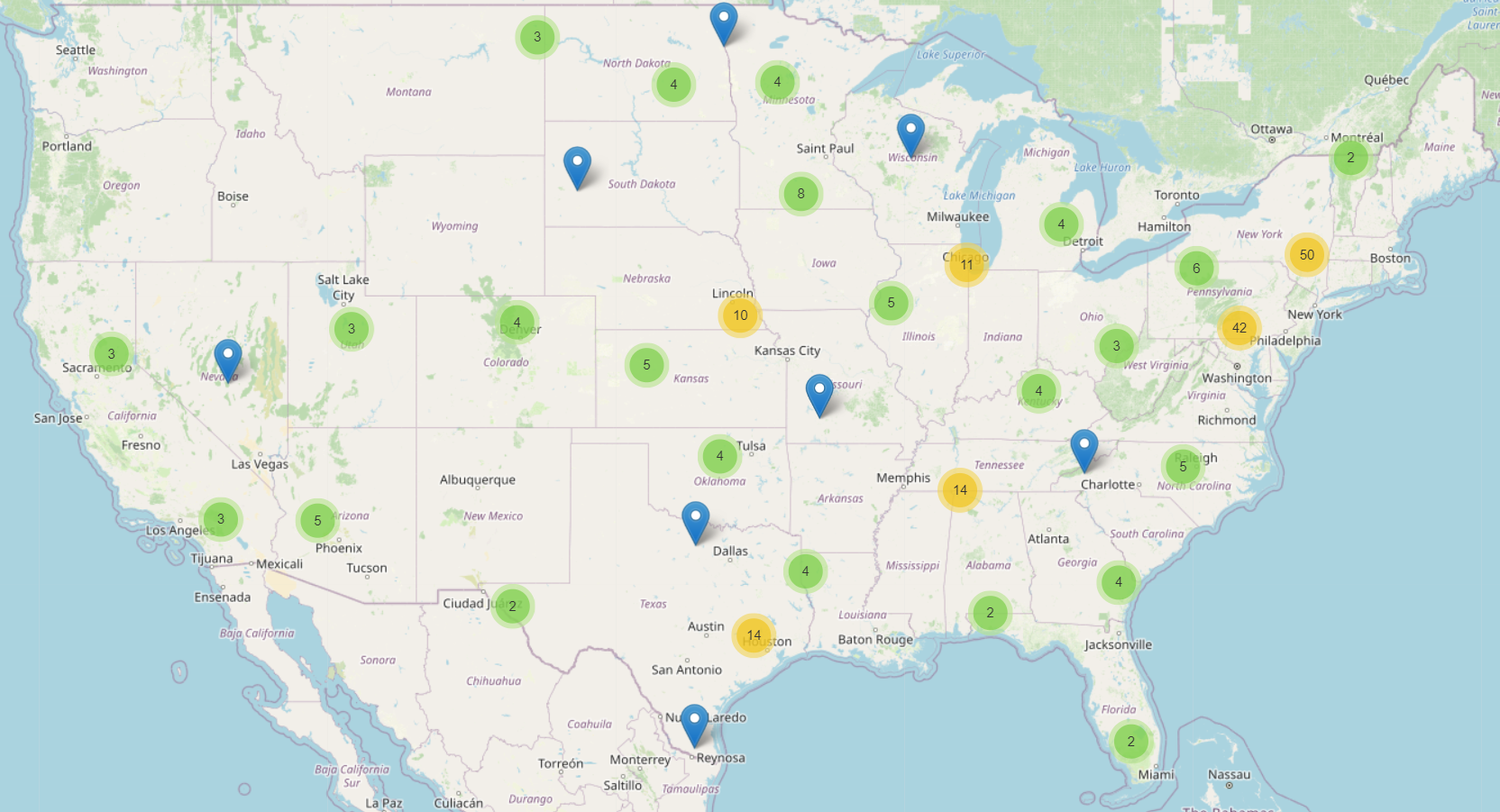}
\centering
\caption{Geographic Distribution of Covered Counties in the Time Series Dataset (The number in the circle stands for the aggregation of nearby counties).}
\label{Fig:geo_vis}
\end{figure*}

The geographic distribution of 238 selected counties in the United States of America is shown in Figure~\ref{Fig:geo_vis}, where the circle with numbers denotes the aggregation of spatially near counties. Of 238 selected counties, 100 are selected for the top-ranked counties based on the yearly frequency of thunderstorms. The rest are selected randomly and try to provide extra information (e.g., causality discovery).

\subsection{Abnormal Patterns of the Time Series Data}

\begin{table}[h]
\caption{Statistics of Anomaly Weather Patterns (i.e., Thunderstorm Winds) Occurrence in 238 Selected Counties in US.}
\centering
\scalebox{1}{
\begin{tabular}{cccccc}
\hline
Year & 2017 & 2018 & 2019 & 2020 & 2021 \\ \hline
Jan  & 26   & 3    & 2    & 41   & 7    \\ 
Feb  & 53   & 6    & 9    & 50   & 8    \\ 
Mar  & 85   & 16   & 26   & 63   & 62   \\ 
Apr  & 93   & 44   & 140  & 170  & 60   \\ 
May  & 245  & 207  & 263  & 175  & 218  \\ 
Jun  & 770  & 302  & 348  & 331  & 452  \\ 
Jul  & 306  & 291  & 457  & 453  & 701  \\ 
Aug  & 294  & 269  & 415  & 354  & 435  \\ 
Sep  & 61   & 80   & 122  & 29   & 123  \\ 
Oct  & 32   & 32   & 82   & 60   & 55   \\ 
Nov  & 20   & 22   & 9    & 114  & 11   \\ 
Dec  & 5    & 15   & 11   & 8    & 58   \\ \hline
\end{tabular}}
\label{tb:label_distribution}
\end{table}

\end{document}